\def\paperID{0000}
\def\confName{CVPR}
\def\confYear{2024}

\newif\ifreview 
\newif\ifarxiv 
\newif\ifcamera \newcommand{\cameraready}{\cameratrue}
\newif\ifrebuttal 

\cameraready

\pdfoutput=1
\documentclass[10pt,twocolumn,letterpaper]{article}
\ifreview \usepackage[review]{cvpr} \fi
\ifarxiv \usepackage[pagenumbers]{cvpr} \fi
\ifrebuttal \usepackage[rebuttal]{cvpr} \fi
\ifcamera \usepackage{cvpr} \fi

\usepackage{graphicx}	
\usepackage{amsmath}	
\usepackage{amssymb}	
\usepackage{booktabs}
\usepackage{times}
\usepackage{microtype}
\usepackage{epsfig}
\usepackage[table,xcdraw,dvipsnames]{xcolor}
\usepackage{caption}
\usepackage{float}
\usepackage{placeins}
\usepackage{color, colortbl}
\usepackage{stfloats}
\usepackage{enumitem}
\usepackage{tabularx}
\usepackage{xstring}
\usepackage{multirow}
\usepackage{xspace}
\usepackage{url}
\usepackage{subcaption}
\usepackage{xcolor}
\usepackage[hang,flushmargin]{footmisc}

\ifcamera \usepackage[accsupp]{axessibility} \fi

% ...

\ifarxiv  \fi

\newcommand{\R}[1]{{%
    \textbf{%
        \ifstrequal{#1}{1}{\textcolor{red}{R#1}}{%
        \ifstrequal{#1}{2}{\textcolor{blue}{R#1}}{%
        \ifstrequal{#1}{3}{\textcolor{magenta}{R#1}}{%
        \ifstrequal{#1}{4}{\textcolor{teal}{R#1}}{%
                           \textcolor{cyan}{R#1}%
        }}}}%
    }%
}}

\usepackage{xr-hyper}

\makeatletter
\newcommand*{\addFileDependency}[1]{
  \typeout{(#1)}
  \@addtofilelist{#1}
  \IfFileExists{#1}{}{\typeout{No file #1.}}
}

\makeatother

\definecolor{cvprblue}{rgb}{0.21,0.49,0.74}
\usepackage[pagebackref,breaklinks,colorlinks,citecolor=cvprblue]{hyperref}
\usepackage[capitalize]{cleveref}
\crefname{section}{Sec.}{Secs.}
\crefname{table}{Table}{Tables}
\crefname{figure}{Fig.}{Figs.}

\usepackage{url}
\usepackage{arydshln}
\usepackage{footnote}
\usepackage{multirow}
\usepackage{pifont}

\usepackage[noEnd=false]{algpseudocodex}
\usepackage{algorithm}

\def\eg{\emph{e.g}\onedot} 
\def\ie{\emph{i.e}\onedot} 

\renewcommand{\thefootnote}{\fnsymbol{footnote}}

\newcommand{\cmark}{\ding{51}}%
\newcommand{\xmark}{\ding{55}}%
\newcommand{\filledstar}{\ding{72}} 

\title{LivePortrait: Efficient Portrait Animation with Stitching and Retargeting Control}

\author{
    \normalsize Jianzhu Guo$^{1 * \dag }$ \quad Dingyun Zhang$^{1,2*}$ \quad Xiaoqiang Liu$^{1}$ \quad Zhizhou Zhong$^{1,3}$ \quad Yuan Zhang$^{1}$ \quad Pengfei Wan$^{1}$ \quad Di Zhang$^{1}$ \\
    \small $^{1}$Kuaishou Technology \quad $^{2}$University of Science and Technology of China \quad $^{3}$Fudan University \\
    \tt\small \url{https://liveportrait.github.io}
}

\begin{document}

\twocolumn[{
    \renewcommand\twocolumn[1][]{#1}
    \maketitle
    \vspace*{-2.9em}
    \begin{center}
        \captionsetup{type=figure}
        \includegraphics[width=0.912\linewidth]{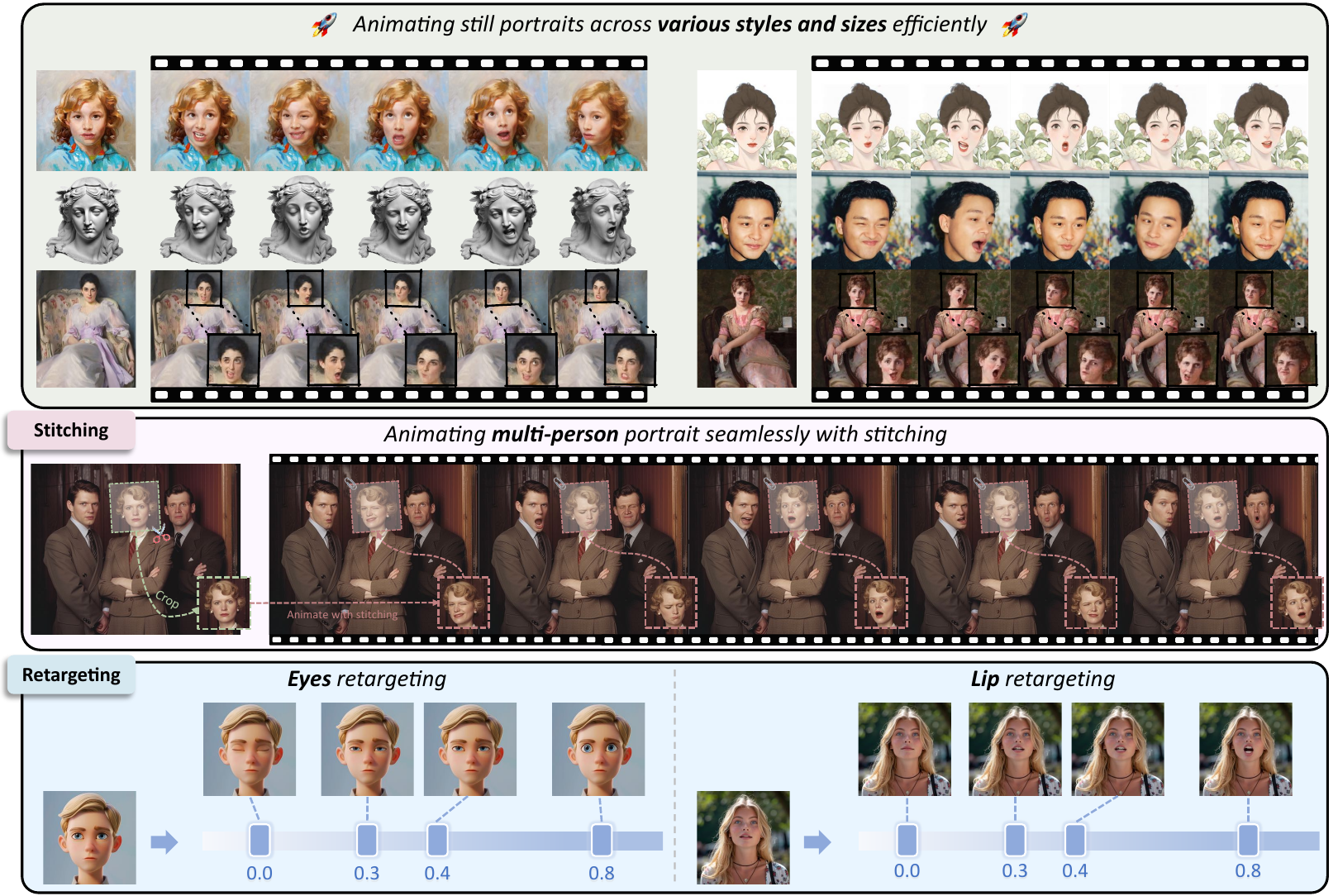} \vspace{-0.8em}
        \captionof{figure}{
            \textbf{Qualitative portrait animation results from our model.} Given a static portrait image as input, our model can vividly animate it, ensuring seamless stitching and offering precise control over eyes and lip movements.
        }
        \label{fig:expressiveness}
    \end{center}
}]

\footnotetext{$^*$Equal contributions. $^\dag$Corresponding author.}
\renewcommand{\thefootnote}{\arabic{footnote}} 

\begin{abstract}
Portrait animation aims to synthesize a lifelike video from a single source image, using it as an appearance reference, with motion (i.e., facial expressions and head pose) derived from a driving video, audio, text, or generation.
Instead of following mainstream diffusion-based methods, we explore and extend the potential of the implicit-keypoint-based framework, which effectively balances computational efficiency and controllability.
Building upon this, we develop a video-driven portrait animation framework named \textbf{LivePortrait} with a focus on better generalization, controllability, and efficiency for practical usage.
To enhance the generation quality and generalization ability, we scale up the training data to about 69 million high-quality frames, adopt a mixed image-video training strategy, upgrade the network architecture, and design better motion transformation and optimization objectives.
Additionally, we discover that compact implicit keypoints can effectively represent a kind of blendshapes and meticulously propose a stitching and two retargeting modules, which utilize a small MLP with negligible computational overhead, to enhance the controllability.
Experimental results demonstrate the efficacy of our framework even compared to diffusion-based methods.
The generation speed remarkably reaches 12.8ms on an RTX 4090 GPU with PyTorch.
The inference code and models are available at \url{https://github.com/KwaiVGI/LivePortrait}.

\end{abstract}

\section{Introduction}
\label{sec:intro}
Nowadays, people frequently use smartphones or other recording devices to capture static portraits to record their precious moments.
The Live Photos\footnote{\url{https://support.apple.com/en-sg/104966}} feature on iPhone can bring static portraits to life by recording the moments $1.5$ seconds before and after a picture is taken, which is likely achieved through a form of video recording.
However, based on recent advances like GANs~\cite{goodfellow2014generative} and Diffusions~\cite{rombach2022high,ho2020denoising,song2020denoising}, various portrait animation methods~\cite{wang2021facevid2vid,hong2022depth,zhao2022thin,hong2023implicit,zeng2023face,mallya2022implicit,siarohin2019first,wei2024aniportrait,xie2024x} have made it possible to animate a static portrait into dynamic ones, without relying on specific recording devices.

In this paper, we aim to animate a static portrait image, making it realistic and expressive, while also pursuing high inference efficiency and precise controllability.
Although diffusion-based portrait animation methods~\cite{wei2024aniportrait,xie2024x,ma2024followyouremoji} have achieved impressive results in terms of quality, they are usually computationally expensive and lack the precise controllability, \eg, stitching control\footnote{\url{https://www.d-id.com/liveportrait-4}}.
Instead, we extensively explore implicit-keypoint-based video-driven frameworks~\cite{siarohin2019first,wang2021facevid2vid}, and extend their potential to effectively balance the generalization ability, computational efficiency, and controllability.

Specifically, we first enhance a powerful implicit-keypoint-based method~\cite{wang2021facevid2vid}, by scaling up the training data to about 69 million high-quality portrait images, introducing a mixed image-video training strategy, upgrading the network architecture, using the scalable motion transformation, designing the landmark-guided implicit keypoints optimization and several cascaded loss terms.
Additionally, we discover that compact implicit keypoints can effectively represent a kind of implicit blendshapes, and meticulously design a stitching module and two retargeting modules, which utilize a small MLP and add negligible computational overhead, to enhance the controllability, such as stitching control.
Our core contributions can be summarized as follows:
(i) developing a solid implicit-keypoint-based video-driven portrait animation framework that significantly enhances the generation quality and generalization ability, and
(ii) designing an advanced stitching module and two retargeting modules for better controllability, with negligible computational overhead.
Extensive experimental results demonstrate the efficacy of our framework, even compared to heavy diffusion-based methods.
Besides, our model can generate a portrait animation in 12.8ms on an RTX 4090 GPU using PyTorch for inference.

\section{Related Work}
Recent video-driven portrait animation methods can be divided into non-diffusion-based and diffusion-based methods, as summarized in \cref{tab:related work}.
\label{sec:related}

\begin{table*}[htbp]
    \centering
    \resizebox{0.95\linewidth}{!}{
    \begin{tabular}{lccccccc}
    \toprule[1.5pt]
    \multirow{2}[2]{*}{\textbf{Method}} & \multirow{2}[2]{*}{\textbf{Framework}} & \multirow{2}[2]{*}{\textbf{Intermediate motion representation}} & \multirow{2}[2]{*}{\textbf{Generation ability}} & \multirow{2}[2]{*}{\textbf{Inference efficiency}} & \multicolumn{3}{c}{\textbf{Controllability}} \\ \cmidrule[0.5pt](lr){6-8}
    & & & & & Stitching & Eyes retargeting & Lip retargeting \\
        \midrule[1pt]
        \noalign{\vskip 0.5em} 
        \noalign{\vskip 0.5em} 
        \begin{tabular}[c]{@{}l@{}}
            FOMM~\cite{siarohin2019first} \\
            MRAA~\cite{siarohin2021motion} \\
            Face Vid2vid~\cite{wang2021facevid2vid} \\
            IWA~\cite{mallya2022implicit} \\
            TPSM~\cite{zhao2022thin} \\
            DaGAN~\cite{hong2022depth} \\
            MCNet~\cite{hong2023implicit} \\
        \end{tabular} & Non-diffusion & Implicit keypoints & \filledstar \filledstar \filledstar & \includegraphics[height=1em]{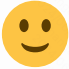} & \xmark &  \xmark & \xmark\\
        \noalign{\vskip 0.5em} 

        \hline
        \noalign{\vskip 0.5em} 
        FADM~\cite{zeng2023face} & Diffusion & Implicit keypoints $+$ 3DMMs & \filledstar \filledstar \filledstar & \includegraphics[height=1em]{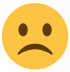} & \xmark &  \xmark & \xmark \\
        \noalign{\vskip 0.5em} 

        \hdashline
        \noalign{\vskip 0.5em} 
        \begin{tabular}[c]{@{}l@{}}
        Face Adapter~\cite{han2024face} \\
        AniPortrait~\cite{wei2024aniportrait} \\
        \end{tabular} & Diffusion & Explicit keypoints or masks & \filledstar \filledstar \filledstar \filledstar& \includegraphics[height=1em]{./figs/frownie.png} & \xmark &  \xmark & \xmark\\
        \noalign{\vskip 0.5em} 
        \hdashline
        \noalign{\vskip 0.5em} 
        \begin{tabular}[c]{@{}l@{}}
        X-Portrait~\cite{xie2024x} \\
        \end{tabular} & Diffusion & Only original driving images & \filledstar \filledstar \filledstar \filledstar \filledstar& \includegraphics[height=1em]{./figs/frownie.png} & \xmark &  \xmark & \xmark\\
        \noalign{\vskip 0.5em} 

        \hdashline
        \noalign{\vskip 0.5em} 
        \begin{tabular}[c]{@{}l@{}}
            MegActor~\cite{yang2024megactor}
        \end{tabular} & Diffusion & Only original driving images & \filledstar \filledstar \filledstar \filledstar& \includegraphics[height=1em]{./figs/frownie.png} & \xmark &  \xmark & \xmark\\
        \noalign{\vskip 0.5em} 

        \midrule[1pt]
        \textbf{Ours} & Non-diffusion & Implicit keypoints & \filledstar \filledstar \filledstar \filledstar & \includegraphics[height=1em]{./figs/smiley.png} & \cmark & \cmark & \cmark \\ 
    \bottomrule[1.5pt]
    \end{tabular}}
    \caption{\textbf{Summary of the video-driven portrait animation methods.}}
    \label{tab:related work}
\end{table*}
\subsection{Non-diffusion-based Portrait Animation}
%
%
%
For non-diffusion-based models, the implicit-keypoints-based methods employed implicit keypoints as the intermediate motion representation, and warped the source portrait with the driving image by the optical flow.
FOMM~\cite{siarohin2019first} performed first-order Taylor expansion near each keypoint and approximated the motion in the neighborhood of
each keypoint using local affine transformations.
MRAA~\cite{siarohin2021motion} represented articulated motion with PCA-based motion estimation.
Face vid2vid~\cite{wang2021facevid2vid} extended FOMM by introducing 3D implicit keypoints representation and achieved free-view portrait animation.
IWA~\cite{mallya2022implicit} improved the warping mechanism based on cross-modal attention, which can be extended to using multiple source images.
To estimate the optical flow more flexibly and work better for large-scale motions, TPSM~\cite{zhao2022thin} used nonlinear thin-plate spline transformation for representing more complex motions.
Simultaneously, DaGAN~\cite{hong2022depth} leveraged the dense depth maps to estimate implicit keypoints that capture the critical driving movements.
MCNet~\cite{hong2023implicit} designed an identity representation conditioned memory compensation network to tackle the ambiguous generation caused by the complex driving motions.

Several works~\cite{khakhulin2022realistic,tewari2020stylerig,ghosh2020gif} employed predefined motion representations, such as 3DMM blendshapes~\cite{blanz2023morphable}.
Another line of works~\cite{megaportraits,drobyshev2024emoportraits} proposed to learn the latent expression representation from scratch.
MegaPortrait~\cite{megaportraits} used the high-resolution images beyond the medium-resolution training images to upgrade animated resolution to megapixel.
EMOPortraits~\cite{drobyshev2024emoportraits} employed an expression-riched training video dataset and the expression-enhanced loss to express the intense motions.

\subsection{Diffusion-based Portrait Animation}
Diffusion models~\cite{rombach2022high,ho2020denoising,song2020denoising} synthesized the desired data samples from Gaussian noise via removing noises iteratively.
\cite{rombach2022high} proposed the Latent Diffusion Models (LDMs) and transferred the training and inference processes to a compressed latent space for efficient computing.
LDMs have been broadly applied to many concurrent works in full-body dance generation~\cite{hu2023animate,chang2024magicpose,xu2023magicanimate,karras2023dreampose,wang2023disco}, audio-driven portrait animation~\cite{qi2023difftalker,shen2023difftalk,tian2024emo,xu2024vasa,wei2024aniportrait,liu2024anitalker,wang2024V-Express}, and video-driven portrait animation~\cite{zeng2023face,wei2024aniportrait, xie2024x, han2024face}.
%

FADM~\cite{zeng2023face} was the first diffusion-based portrait animation method.
It obtained the coarsely animated result via the pretrained implicit-keypoints-based model and then got the final animation under the guidance of the 3DMMs with the diffusion model.
%
Face Adapter~\cite{han2024face} used an identity adapter to enhance the identity preservation of the source portrait and a spatial condition generator to generate the explicit spatial condition, \ie, keypoints and foreground masks, as the intermediate motion representation.
Several works~\cite{wei2024aniportrait,xie2024x,yang2024megactor} employed the mutual self-attention and plugged temporal attention architecture similar to AnimateAnyone~\cite{hu2023animate} to achieve better image quality and appearance preservation.
AniPortrait~\cite{wei2024aniportrait} used the explicit spatial condition, \ie, keypoints, as the intermediate motion representation.
X-Portrait~\cite{xie2024x} proposed to animate the portraits directly with the original driving video instead of using the intermediate motion representations.
It employed the implicit-keypoint-based method~\cite{wang2021facevid2vid} for cross-identity training to achieve this.
%
MegActor~\cite{yang2024megactor} also animated the source portrait with the original driving video.
It employed the existing face-swapping and stylization framework to get the cross-identity training pairs and encoded the background appearance to improve the animation stability.

\section{Methodology}
\label{sec:method}

This section details our method.
We begin with a brief review of the video-based portrait animation framework face vid2vid~\cite{wang2021facevid2vid} and introduce our significant enhancements aimed at enhancing the generalization ability and expressiveness of animation.
Then, we present our meticulously designed stitching and retargeting modules which provide desired controllability with negligible computational overhead.
Finally, we detail the inference pipeline.

\subsection{Preliminary of Face Vid2vid}
\label{subsec:preliminary}
Face vid2vid~\cite{wang2021facevid2vid} is a seminal framework for animating a still portrait, using the motion features extracted from the driving video sequence.
The original framework consists of an appearance feature extractor $\mathcal{F}$, a canonical implicit keypoint detector $\mathcal{L}$, a head pose estimation network $\mathcal{H}$, an expression deformation estimation network $\Delta$, a warping field estimator $\mathcal{W}$, and a generator $\mathcal{G}$.
$\mathcal{F}$ maps the source image $s$ to a 3D appearance feature volume $f_s$.
The source 3D keypoints $x_{s}$ and the driving 3D keypoints $x_{d}$ are transformed as follows:
\begin{equation}
    \left\{
        \begin{array}{lr}
        x_{s} = x_{c,s} R_{s} + \delta_{s} + t_{s}, &  \\
        x_{d} = x_{c,s} R_{d} + \delta_{d} + t_{d}, &
        \end{array}
    \right.
    \label{eq:k3d_ori}
\end{equation}
where $x_s$ and $x_d$ are the source and driving 3D implicit keypoints, respectively, and $x_{c,s} \in \mathbb{R}^{K \times 3}$ represents the canonical keypoints of the source image. The source and driving poses are $R_s$ and $R_d \in \mathbb{R}^{3 \times 3}$, the expression deformations are $\delta_s$ and $\delta_d \in \mathbb{R}^{K \times 3}$, and the translations are $t_s$ and $t_d \in \mathbb{R}^3$.
Next, $\mathcal{W}$ generates a warping field using the implicit keypoint representations $x_s$ and $x_d$, and employs this flow field to warp the source feature volume $f_s$.
Subsequently, the warped features pass through a decoder generator $\mathcal{G}$, translating them into image space and resulting in a target image.

\subsection{Stage I: Base Model Training}
\label{subsec:stagei_basemodel}

\begin{figure*}[htpb]
    \centering
    \includegraphics[width=0.75\linewidth, trim=0pt 0pt 0pt 0pt, clip]{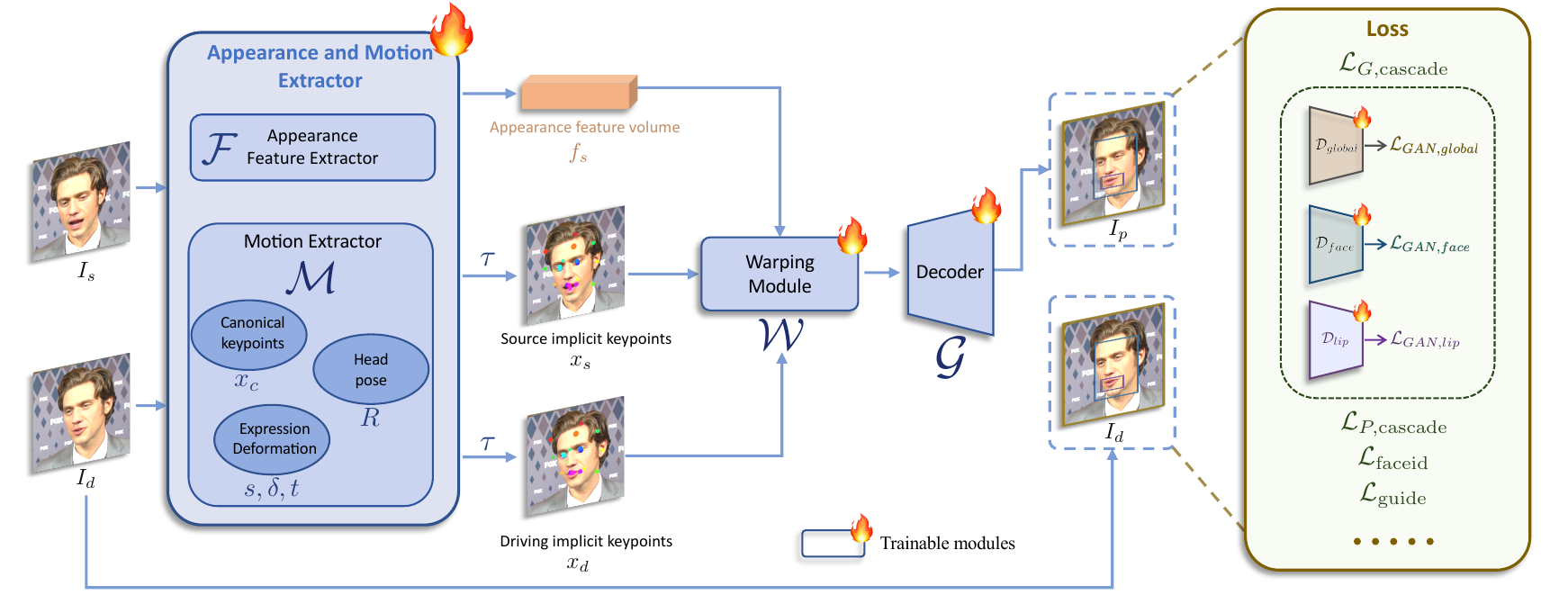} 
    \captionsetup{font=scriptsize}
    \caption{
        \textbf{Pipeline of the first stage: base model training.}
        The appearance and motion extractors $\mathcal{F}$ and $\mathcal{M}$, the warping module $\mathcal{W}$, and the decoder $\mathcal{G}$ are optimized. In this stage, models are trained from scratch. Please refer to \cref{subsec:stagei_basemodel} for details.
    }
    \label{fig:pipeline_first_stage}
\end{figure*}


We choose face vid2vid~\cite{wang2021facevid2vid} as our base model and introduce a series of significant enhancements.
These include high-quality data curation, a mixed image and video training strategy, an upgraded network architecture, scalable motion transformation, landmark-guided implicit keypoints optimization, and cascaded loss terms.
These advancements significantly enhance the expressiveness of the animation and the generalization ability of the model. The pipeline of the first training stage is shown in \cref{fig:pipeline_first_stage}.

\paragraph{High quality data curation.}
We leverage public video datasets such as Voxceleb~\cite{nagrani2017voxceleb}, MEAD~\cite{kaisiyuan2020mead}, and RAVDESS~\cite{livingstone2018ryerson}, as well as the styled image dataset AAHQ~\cite{liu2021blendgan}. Additionally, we collect a large corpus of 4K-resolution portrait videos with various poses and expressions, 200 hours of talking head videos, and utilize the private LightStage~\cite{yang2023towards,yang2024vrmm} dataset, along with several styled portrait videos and images.
We split long videos into clips of less than 30 seconds and ensure each clip contains only one person using face tracking and recognition.
To maintain the quality of the training data, we use KVQ~\cite{zhao2023quality} to filter out low-quality video clips.
Finally, our training data consists of 69M video frames (92M before filtering) from about 18.9K identities and 60K static styled portraits.

\paragraph{Mixed image and video training.}
The model trained only on realistic portrait videos performs well on human portraits but generalizes poorly to styled portraits, \eg, anime.
Styled portrait videos are scarce, we collect only about 1.3K clips from fewer than 100 identities.
In contrast, high-quality styled portrait images are more abundant; we gathered approximately 60K images, each representing a unique identity, offering diverse identity information.
To leverage both data types, we treat single images as one-frame video clips and train the model on both images and videos.
This mixed training improves the model's generalization ability.

\paragraph{Upgraded network architecture.}
We unify the original canonical implicit keypoint detector $\mathcal{L}$, head pose estimation network $\mathcal{H}$, and expression deformation estimation network $\Delta$ into a single model $\mathcal{M}$, with ConvNeXt-V2-Tiny~\cite{woo2023convnext} as the backbone, which directly predicts the canonical keypoints, head pose and expression deformation of the input image.
Additionally, we follow~\cite{facevid2vid3} to use SPADE decoder~\cite{park2019SPADE} as the generator $\mathcal{G}$, which is more powerful than the original decoder in face vid2vid~\cite{wang2021facevid2vid}.
The warped feature volume $f_s$ is delicately fed into the SPADE decoder, where each channel of the feature volume serves as a semantic map to generate the animated image.
For efficiency, we insert a PixelShuffle~\cite{shi2016real} layer as the final layer of $\mathcal{G}$ to upsample the resolution from $256$$\times$$256$ to $512$$\times$$512$.

\paragraph{Scalable motion transformation.}
The original implicit keypoint transformation in Eqn.~\ref{eq:k3d_ori} ignores the scale factor, which tends to incorporate scaling into the expression deformation and increases the training difficulty.
To address this issue, we introduce a scale factor to the motion transformation, and the updated transformation $\tau$ is formulated as:
\begin{equation}
    \left\{
        \begin{array}{lr}
        x_{s} = s_{s} \cdot ( x_{c,s} R_{s} + \delta_{s}) + t_{s}, &  \\
        x_{d} = s_{d} \cdot ( x_{c,s} R_{d} + \delta_{d}) + t_{d}, &
        \end{array}
    \right.
    \label{eq:k3d_new}
\end{equation}
where $s_{s}$ and $s_{d}$ are the scale factors of the source and driving input, respectively.
Note that the transformation differs from the scale orthographic projection, which is formulated as $x = s \cdot \big( (x_c + \delta) R \big) + t$.
We find that the scale orthographic projection leads to overly flexible learned expressions $\delta$, causing texture flickering when driving across different identities.
Therefore, this transformation can be seen as a tradeoff between flexibility and drivability.

\paragraph{Landmark-guided implicit keypoints optimization.}
The original face vid2vid~\cite{wang2021facevid2vid, facevid2vid3} seems to lack the ability to vividly drive facial expressions, such as winking and eye movements.
In particular, the eye gazes of generated portraits are bound to the head pose and remain parallel to it, a limitation we also observed in our reproduction experiments.
We attribute these limitations to the difficulty of learning subtle facial expressions, like eye movements, in an unsupervised manner.
To address this, we introduce 2D landmarks that capture micro-expressions, using them as guidance to optimize the learning of implicit points.
The landmark-guided loss $\mathcal{L}_{\textrm{guide}}$ is formulated as follows:
\begin{equation}
    \mathcal{L}_{\textrm{guide}} \! = \! \frac{1}{2N} \sum_{i=1}^{N} \big( \textrm{Wing} (l_i, x_{s,i,:2}) \! + \! \textrm{Wing} (l_i, x_{d,i,:2}) \big),
\end{equation}
where $N$ is the number of selected landmarks, $l_i$ is the $i$-th landmark, $x_{s,i,:2}$ and $x_{d,i,:2}$ represent the first two dimensions of the corresponding implicit keypoints respectively, and Wing loss is adopted following~\cite{feng2018wing}.
In our experiments, $N$ is set to 10, with the selected landmarks taken from the eyes and lip.

\paragraph{Cascaded loss terms.}
We follow face vid2vid~\cite{wang2021facevid2vid} to use implicit keypoints equivariance loss $\mathcal{L}_E$, keypoint prior loss $\mathcal{L}_{L}$, head pose loss $\mathcal{L}_{H}$, and deformation prior loss $\mathcal{L}_{\Delta}$.
To further improve the texture quality, we apply perceptual and GAN losses on the global region of the input image, and local regions of face and lip, denoted as a cascaded perceptual loss $\mathcal{L}_{P, \textrm{cascade}}$ and a cascaded GAN loss $\mathcal{L}_{G, \textrm{cascade}}$.
$\mathcal{L}_{G, \textrm{cascade}}$ consists of $\mathcal{L}_{GAN, global}$, $\mathcal{L}_{GAN, face}$, and $\mathcal{L}_{GAN, lip}$, which depend on the corresponding discriminators $\mathcal{D}_{global}$, $\mathcal{D}_{face}$, and $\mathcal{D}_{lip}$ training from scratch.
The face and lip regions are defined by 2D semantic landmarks.
We also adopt a face-id~\cite{deng2018arcface} loss $\mathcal{L}_{\textrm{faceid}}$ to preserve the identity of the source image.
The overall training objective of the first stage is formulated as:
\begin{equation}
    \begin{aligned}
    \mathcal{L}_{\textrm{base}} = &
        \ \mathcal{L}_{E} + \mathcal{L}_{L} + \mathcal{L}_{H} + \mathcal{L}_{\Delta} + \\
        & \ \mathcal{L}_{P, \textrm{cascade}} +
        \mathcal{L}_{G, \textrm{cascade}} +
        \mathcal{L}_{\textrm{faceid}} + \mathcal{L}_{\textrm{guide}}.
    \end{aligned}
\end{equation}
During the first stage, the model is fully trained from scratch.

\subsection{Stage II: Stitching and Retargeting}
\label{subsec:stitching_and_retargeting}

\begin{figure*}[htpb]
    \centering
    \includegraphics[width=1\linewidth, trim=0pt 2pt 0pt 2pt, clip]{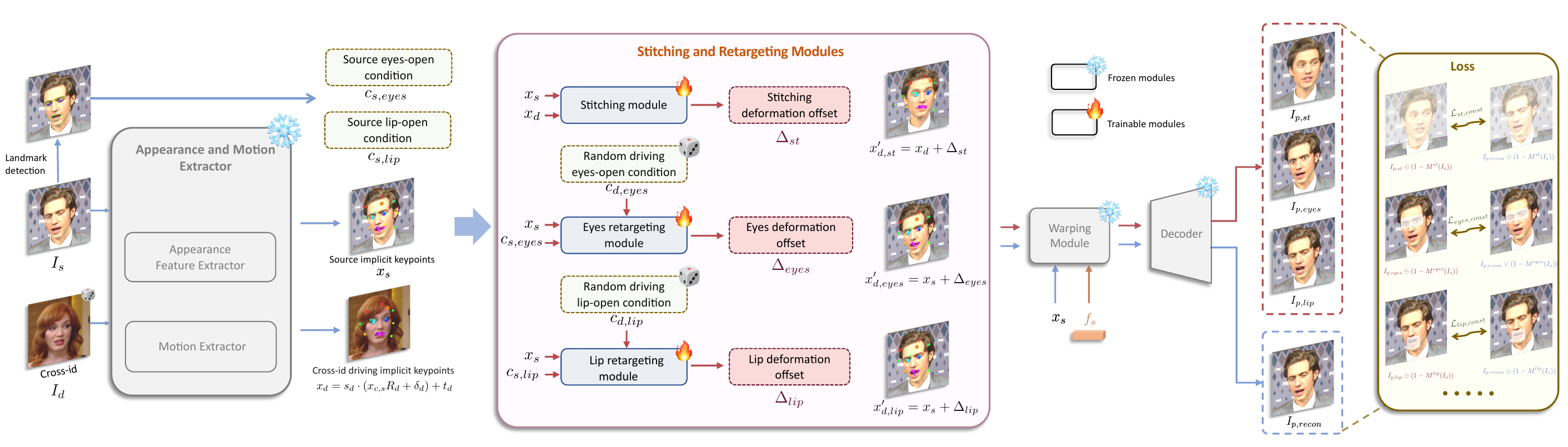} 
    \captionsetup{font=scriptsize}
    \caption{
        \textbf{Pipeline of the second stage: stitching and retargeting modules training.}
        After training the base model in the first stage, we freeze the appearance and motion extractor, warpping module and decoder.
        Only the stitching module and the retargeting modules are optimized in the second stage.
        Please refer to \cref{subsec:stitching_and_retargeting} for details.
    }
    \label{fig:pipeline_second_stage}
\end{figure*}

We suppose the compact implicit keypoints can serve as a kind of implicit blendshapes.
Unlike pose, we cannot explicitly control the expressions, but rather need a combination of these implicit blendshapes to achieve the desired effects.
Surprisingly, we discover that such a combination can be well learned using only a small MLP network, with negligible computational overhead.
Considering practical requirements, we design a stitching module, an eyes retargeting module, and a lip retargeting module.
The stitching module pastes the animated portrait back into the original image space without pixel misalignment, such as in the shoulder region.
This enables the handling of much larger image sizes and the animation of multiple faces simultaneously.
The eyes retargeting module is designed to address the issue of incomplete eye closure during cross-id reenactment, especially when a person with small eyes drives a person with larger eyes.
The lip retargeting module is designed similarly to the eye retargeting module, and can also normalize the input by ensuring that the lips are in a closed state, which facilitates better animation driving.
The pipeline of the second training stage is shown in \cref{fig:pipeline_second_stage}.

\paragraph{Stitching module.}
During training, the stitching module $\mathcal{S}$ receives the source and driving implicit keypoints $x_{s}$ and $x_{d}$ as input, and estimates a deformation offset $\Delta_{st} \in \mathbb{R}^{K \times 3}$ of the driving keypoints.
Following Eqn.~\ref{eq:k3d_new}, the source implicit keypoints are calculated as $x_s = s_{s} \cdot (x_{c,s} R_{s} + \delta_{s}) + t_{s}$, and the driving implicit keypoints $x_{d}$ are calculated using another person's motions as $x_d = s_{d} \cdot (x_{c,s} R_{d} + \delta_{d}) + t_{d}$.
Note that the transformation of $x_d$ differs from the first training stage, as we deliberately use cross-id rather than same-id motion to increase training difficulty, aiming for better generalization in stitching.
Then, $\Delta_{st} = \mathcal{S}(x_s, x_d)$, the driving keypoints are updated as $x'_{d,st} = x_{d} + \Delta_{st}$, and the prediction image $I_{p,st} = \mathcal{D} \big(\mathcal{W}(f_s; x_s, x'_{d, st}) \big)$.
We denote the self-reconstruction image as $I_{p,recon} = \mathcal{D} \big(\mathcal{W}(f_s; x_s, x_s) \big)$.
Finally, the stitching objective $\mathcal{L}_{\textrm{st}}$ is formulated as:
\begin{equation}
    \scalebox{0.9}{$
    \mathcal{L}_{\textrm{st}} = \underbrace{\big\Vert (I_{p,st} \! - \! I_{p,recon}) \odot \big(1 \! - \! M^{st}(I_{s})\big) \big\Vert_1}_{\mathcal{L}_{st,const}} \! +  w_{reg}^{st} \big\Vert \Delta_{st} \big\Vert_1,
    $}
\end{equation}
where $\mathcal{L}_{st,const}$ is the consistency pixel loss between the shoulder region of the prediction and the self-reconstruction image,
$M^{st}$ is a mask operator that masks out the non-shoulder region from the source image $I_{s}$, which is visualized in \cref{fig:pipeline_second_stage}.
$\Vert \Delta_{st} \Vert_1$ is the $L_1$ norm regularization of the stitching deformation offset, and $w_{reg}^{st}$ is a hyperparameter.

\paragraph{Eyes and lip retargeting modules.}
The eyes retargeting module $\mathcal{R}_{eyes}$ receives the source implicit keypoints $x_s$, the source eyes-open condition tuple $c_{s,eyes}$ and a random driving eyes-open scalar $c_{d,eyes} \in [0, 0.8]$ as input, estimating a deformation offset $\Delta_{eyes} \in \mathbb{R}^{K \times 3}$ for the driving keypoints: $\Delta_{eyes} = \mathcal{R}_{eyes}(x_s; c_{s,eyes}, c_{d, eyes})$.
The eyes-open condition denotes the ratio of eye-opening: the larger the value, the more open the eyes.
Similarly, the lip retargeting module $\mathcal{R}_{lip}$ also receives the source implicit keypoints $x_s$ and the source lip-open condition scalar $c_{s,lip}$ and a random driving lip-open scalar $c_{d,lip}$ as input, estimating a deformation offset $\Delta_{lip} \in \mathbb{R}^{K \times 3}$ for the driving keypoints: $\Delta_{lip} = \mathcal{R}_{lip}(x_s; c_{s,lip}, c_{d,lip})$.
Then, the driving keypoints are updated as $x'_{d,eyes} = x_{s} + \Delta_{eyes}$ and $x'_{d,lip} = x_{s} + \Delta_{lip}$, and the prediction images are $I_{p,eyes} = \mathcal{D} \big(\mathcal{W}(f_s; x_s, x'_{d, eyes}) \big)$ and $I_{p,lip} = \mathcal{D} \big(\mathcal{W}(f_s; x_s, x'_{d, lip}) \big)$.
Finally, the training objectives for the eyes and lip retargeting modules are formulated as follows:
\begin{equation}
    \scalebox{0.8}{$
    \begin{aligned}
    \mathcal{L}_{\textrm{eyes}} = &
        \ \underbrace{\big\Vert (I_{p,eyes} \! - \! I_{p,recon}) \odot \big(1 \! - \! M^{eyes}(I_{s})\big) \big\Vert_1}_{\mathcal{L}_{eyes,const}}  +  \\
        & w_{cond}^{eyes} \big\Vert c^{p}_{s,eyes} - c_{d,eyes} \big\Vert_1 + w_{reg}^{eyes} \big\Vert \Delta_{eyes} \big\Vert_1, \\ \\
    \mathcal{L}_{\textrm{lip}} = &
        \ \underbrace{\big\Vert (I_{p,lip} \! - \! I_{p,recon}) \odot \big(1 \! - \! M^{lip}(I_{s})\big) \big\Vert_1}_{\mathcal{L}_{lip,const}} + \\
        & w_{cond}^{lip} \big\Vert c^{p}_{s,lip} - c_{d,lip} \big\Vert_1 +  w_{reg}^{lip} \big\Vert \Delta_{lip} \big\Vert_1, \\
    \end{aligned}
    $}
\end{equation}
where $M^{eyes}$ and $M^{lip}$ are mask operators that mask out the eyes and lip regions from the source image $I_{s}$ respectively, $c^{p}_{s,eyes}$ and $c^{p}_{s,lip}$ are the condition tuples from $I_{p,eyes}, I_{p,lip}$ respectively, and $w_{cond}^{eyes}, w_{reg}^{eyes}, w_{cond}^{lip}, w_{reg}^{lip}$ are hyperparameters.


\subsection{Inference}



In the inference phase, we first extract the feature volume $f_s = \mathcal{F}(I_s)$, the canonical keypoints $x_{c,s} = \mathcal{M}(I_s)$ from the source image $I_s$.
Given a driving video sequence $\{ {I_{d,i} | i=0, \ldots, N-1} \}$,
we extract motions from each frame $s_{d,i}, \delta_{d,i}, t_{d,i}, R_{d,i} = \mathcal{M} (I_{d,i})$ and conditions $c_{d,eyes,i}$ and $c_{d,lip,i}$.
The source and driving implicit keypoints are next transformed as follows:
\begin{equation}
    \left\{
        \begin{array}{ll}
        x_{s} &= s_{s} \cdot (x_{c,s} R_{s} + \delta_{s}) + t_{s}, \\
        x_{d,i} &= s_{s} \cdot \frac{s_{d,i}}{s_{d,0}} \cdot \big( x_{c,s} (R_{d,i} R^{-1}_{d,0} R_s) + \\
                        &(\delta_{s} + \delta_{d,i} - \delta_{d,0})\big) + (t_{s} + t_{d,i} - t_{d,0}).
        \end{array}
    \right.
    \label{eqn:inference}
\end{equation}
Then, the influence procedure could be described as in \cref{algo:inference}, where \(\alpha_{st}\), \(\alpha_{eyes}\) and \(\alpha_{lip}\) are indicator variables that can take values of either $0$ or $1$.
The final prediction image $I_{p,i}$ is generated by the warping network $\mathcal{W}$ and the decoder $\mathcal{D}$.
Note that the deformation offsets of eyes and lip are decoupled from each other, allowing them to be linearly added to the driving keypoints.

\begin{algorithm}[htbp]
    \small
    \caption{Illustration of the inference procedure}
    \label{algo:inference}
    \begin{algorithmic}[1]
        \Statex\textbf{Input:} $f_s; x_{s}, x_{d,i}; \alpha_{eyes}, \alpha_{lip}, \alpha_{st}; c_{s,eyes}, c_{d, eyes, i}, c_{s,lip}$,
        \Statex $c_{d, lip, i}$
        \Statex\textbf{Output:} $I_{p,i}$

        \If{$\alpha_{st} = 0 \mathrm{~and~} \alpha_{eyes} = 0 \mathrm{~and~} \alpha_{lip} = 0$}
        \LComment{without stitching or retargeting}

        \State $x'_{d,i} \gets x_{d,i}$

        \ElsIf{$\alpha_{st} = 1 \mathrm{~and~} \alpha_{eyes} = 0 \mathrm{~and~} \alpha_{lip} = 0$}
        \LComment{with stitching and without retargeting}
        \State $\Delta_{st,i} = \mathcal{S}(x_s, x_{d,i})$
        \State $x'_{d,i} \gets x_{d,i} + \Delta_{st,i}$

        \ElsIf{$\alpha_{eyes} = 1 \mathrm{~or~} \alpha_{lip} = 1$}
        \LComment{with eyes or lip retargeting}
        \State $\Delta_{eyes,i} = \mathcal{R}_{eyes}(x_s; c_{s,eyes}, c_{d,eyes,i})$
        \State $\Delta_{lip,i} = \mathcal{R}_{lip}(x_s; c_{s,lip}, c_{d, lip, i})$
        \State $x'_{d,i} \gets x_{s} + \alpha_{eyes}  \Delta_{eyes,i} + \alpha_{lip}  \Delta_{lip,i}$

        \If{$\alpha_{st} = 1$}
        \State $\Delta_{st,i} = \mathcal{S}(x_s, x'_{d,i})$
        \State $x'_{d,i} \gets x'_{d,i} + \Delta_{st,i}$
        \EndIf

        \EndIf

        \State $I_{p,i} \gets \mathcal{D} \big( \mathcal{W} (f_s; x_s, x'_{d,i} ) \big)$

    \end{algorithmic}
\end{algorithm}

\section{Experiments}
\label{sec:exp}
We first give an overview of the implementation details, baselines, and benchmarks used in the experiments.
Then, we present the experimental results on self-reenactment and cross-reenactment, followed by an ablation study to validate the effectiveness of the proposed stitching and retargeting modules.

\paragraph{Implementation Details.}
In the first training stage, our models are trained from scratch using 8 NVIDIA A100 GPUs for approximately 10 days.
During the second training stage, we only train the stitching and retargeting modules while keeping other parameters frozen, which takes approximately 2 days.
The input images are aligned and cropped to a resolution of $256$$\times$$256$, with a batch size set to $104$.
The output resolution is $512$$\times$$512$.
The Adam optimizer is employed with a learning rate of $2\times10^{-4}$, $\beta_1=0.5$, and $\beta_2=0.999$.
The stitching module consists of a four-layer MLP with layer sizes of $[126, 128, 128, 64, 65]$.
The eyes retargeting module consists of a six-layer MLP with layer sizes of $[66, 256, 256, 128, 128, 64, 63]$.
The lip retargeting module consists of a four-layer MLP with layer sizes of $[65, 128, 128, 64, 63]$.
The computation budget of the stitching and retargeting modules is negligible.

%
\paragraph{Baselines.}
We compare our model with several non-diffusion-based methods, including FOMM~\cite{siarohin2019first}, Face Vid2vid~\cite{wang2021facevid2vid}, DaGAN~\cite{hong2022depth}, MCNet~\cite{hong2023implicit}, and TPSM~\cite{zhao2022thin}, as well as diffusion-based models such as FADM~\cite{zeng2023face}, AniPortrait~\cite{wei2024aniportrait}, and X-Portrait~\cite{xie2024x}.
For face vid2vid~\cite{wang2021facevid2vid}, we employ the implementation from~\cite{facevid2vid3}, while for the other methods, we use the official implementations.
\paragraph{Benchmarks.}
To measure the generalization quality and motion accuracy of portrait animation results, we adopt Peak Signal-to-Noise Ratio (PSNR), Structural Similarity Index (SSIM)~\cite{wang2004image}, Learned Perceptual Image Patch Similarity (LPIPS)~\cite{zhang2018unreasonable}, $\mathcal{L}_1$ distance, FID~\cite{heusel2017gans}, Average Expression Distance (AED)~\cite{siarohin2019first}, Average Pose Distance (APD)~\cite{siarohin2019first}, and Mean Angular Error (MAE) of eyeball direction~\cite{han2024face}.
For self-reenactment, our models are evaluated on the official test split of the TalkingHead-1KH dataset~\cite{wang2021facevid2vid} and VFHQ dataset~\cite{xie2022vfhq}, which consist of 35 and 50 videos respectively.
For cross-reenactment, the first 50 images obtained from the FFHQ dataset~\cite{karras2019style} are used as source portraits.
Detailed descriptions of these metrics are provided in \cref{sec:appendix_benchmark}.

\subsection{Self-reenactment}
For each test video sequence, we use the first frame as the source input and animate it using the whole frames as driving images, which also serve as ground truth.
%
For comparisons, the animated portraits and the ground truth images are downsampled to a resolution of $256$$\times$$256$ to maintain consistency with the baselines.
Qualitative and quantitative comparisons are detailed as follows.

\begin{figure*}[htpb]
  \centering
  \includegraphics[width=0.95\linewidth]{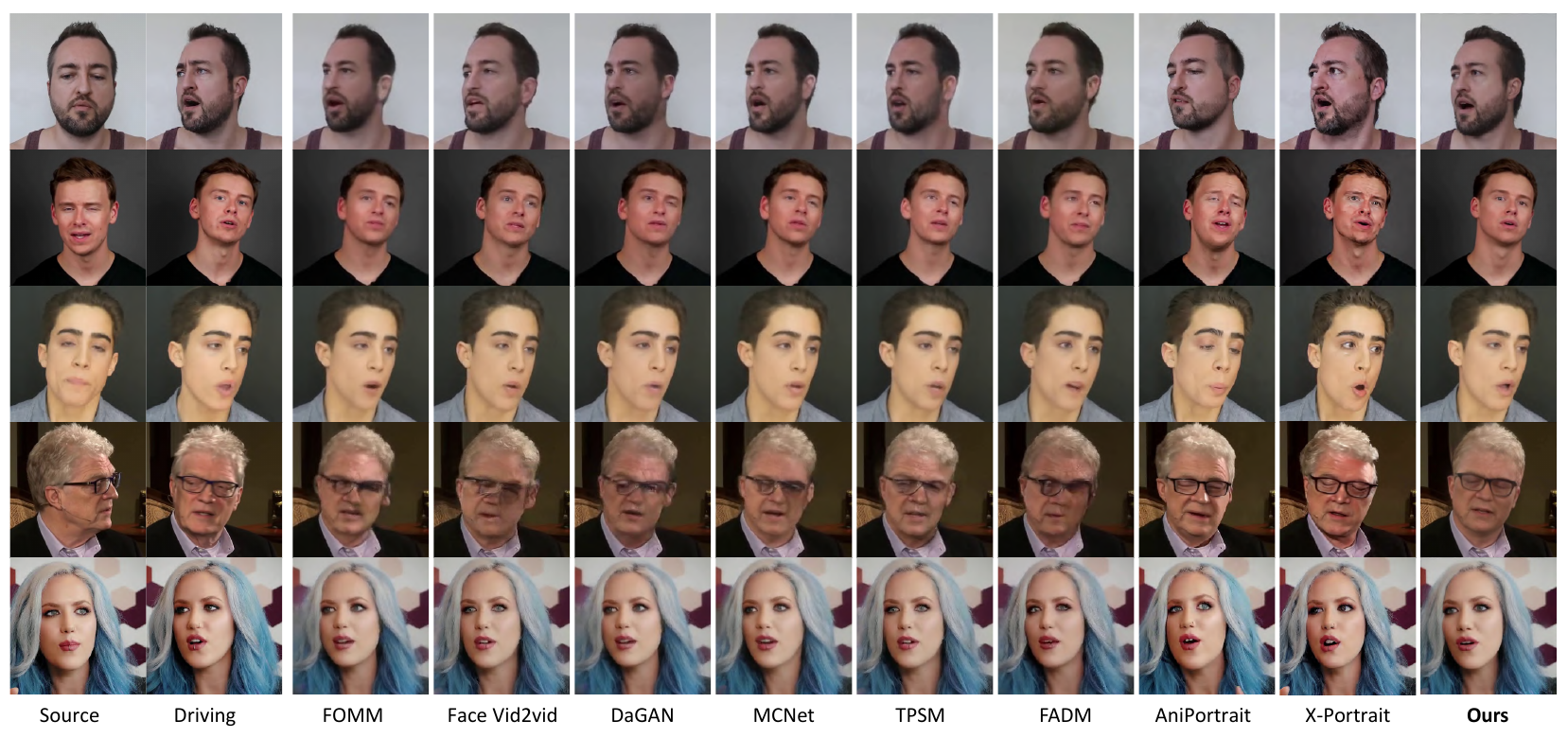}
  \caption{
    \textbf{Qualitative comparisons of self-reenactment.}
    The first four source-driving paired images are from TalkingHead-1KH~\cite{wang2021facevid2vid} and the last ones are from VFHQ~\cite{xie2022vfhq}.
    Our model faithfully preserves lip movements and eye gazes, handles large poses more stably, and maintains the identity of the source portrait better compared to other methods.}
  \label{fig:self-driven}
\end{figure*}

\paragraph{Qualitative results.}
The qualitative comparisons are illustrated in \cref{fig:self-driven}.
Our results are the pasted back images in the original image space using the first-stage base model.
These cases demonstrate that our model can faithfully transfer motions from the driving images, including lip movements and eye gazes, while preserving the appearance details of the source portrait.
%
%
The fourth case in \cref{fig:self-driven} demonstrates that our model achieves stable animation results even with large poses, ensuring accurate transfer of poses.

\paragraph{Quantitative results.}
In \cref{tab:self-driven}, we present quantitative comparisons of self-reenactment.
Our model slightly outperforms previous diffusion-based methods, such as FADM~\cite{zeng2023face}, AniPortrait~\cite{wei2024aniportrait}, and X-Portrait~\cite{xie2024x}, in generation quality, and demonstrates better eyes motion accuracy than other methods.

\begin{table*}[t]
  \centering
  \resizebox{0.95\linewidth}{!}{
  \begin{tabular}{lcccccccccccc}
      \toprule[1.5pt]
      \multirow{2}[2]{*}{Method}  & \multicolumn{6}{c}{\textbf{TalkingHead-1KH}} & \multicolumn{6}{c}{\textbf{VFHQ}} \\ \cmidrule[0.5pt](lr){2-7} \cmidrule[0.5pt](lr){8-13}
       & PSNR$\uparrow$ & SSIM$\uparrow$ & LPIPS$\downarrow$ & $\mathcal{L}_1$$\downarrow$ & CSIM$\uparrow$ & MAE (\textdegree)$\downarrow$ & PSNR$\uparrow$ & SSIM$\uparrow$ & LPIPS$\downarrow$ & $\mathcal{L}_1$$\downarrow$ & CSIM$\uparrow$ & MAE (\textdegree)$\downarrow$ \\
      \midrule[1pt]
       FOMM~\cite{siarohin2019first}  & 31.0681 & 0.7620 & 0.1201 & 0.0419  & 0.8805 & 10.1745 & 30.5912 & 0.7098 & 0.1410 & 0.0505 & 0.8700 & 10.9327 \\
       Face Vid2vid \cite{wang2021facevid2vid,facevid2vid3}  & 30.8438 & 0.7743 & 0.0940 & 0.0432 & 0.8774 & 10.8117 & 30.5166 & 0.7247 & 0.1132 & 0.0500 & 0.8775 & 11.1500 \\
       DaGAN \cite{hong2022depth}  & 31.3657 & 0.7903 & 0.0969 & 0.0389  & 0.8798 & 11.8655 & 30.7038 & 0.7315 & 0.1258 & 0.0481 & 0.8747 & 11.2051 \\
       MCNet~\cite{hong2023implicit} & \underline{32.0013} & \underline{0.8042} & 0.1018 & \underline{0.0349} & \underline{0.8876} & 10.9035 & \underline{31.3459} & \underline{0.7540} & 0.1209 & \underline{0.0429} & 0.8849 & \underline{9.6634} \\
       TPSM~\cite{zhao2022thin} & 31.2934 & 0.7965 & 0.0990 & 0.0395 & 0.8848 & \underline{9.6036} & 31.0262 & 0.7476 & 0.1177 & 0.0466 & \underline{0.8884} & 9.8169 \\
       \midrule[1pt]
       FADM~\cite{zeng2023face} & 30.2141 & 0.7695 & 0.1049 & 0.0484 & 0.8708 & 11.4484 & 30.0932 & 0.7180 & 0.1252 & 0.0535 & 0.8707 & 11.7523 \\
       AniPortrait~\cite{wei2024aniportrait} & 31.4669 & 0.7144 & \underline{0.0922} & 0.0470 & 0.8550 & 12.0807 & 30.9013 & 0.6718 & \underline{0.1073} & 0.0542 & 0.8570 & 14.2411 \\
       X-Portrait~\cite{xie2024x} & 31.2716 & 0.7193 & 0.1007 & 0.0487 & 0.8773 & \underline{9.2335} & 30.5840 & 0.6479 & 0.1312 & 0.0627 & 0.8721 & \underline{9.3846} \\
       \midrule[1pt]
      \textbf{Ours} & \textbf{32.0082} & \textbf{0.8193} & \textbf{0.0664} & \textbf{0.0347} & \textbf{0.9125} & \textbf{7.0535} & \textbf{31.5616} & \textbf{0.7653} & \textbf{0.0798} & \textbf{0.0422} & \textbf{0.9121} & \textbf{6.6966} \\
      \bottomrule[1.5pt]
  \end{tabular}
  }
  \caption{\textbf{Quantitative comparisons of self-reenactment.}
  }
  \label{tab:self-driven}
\end{table*}

%
%

\subsection{Cross-reenactment}
\begin{figure*}[htpb]
  \centering
  \includegraphics[width=0.95\linewidth]{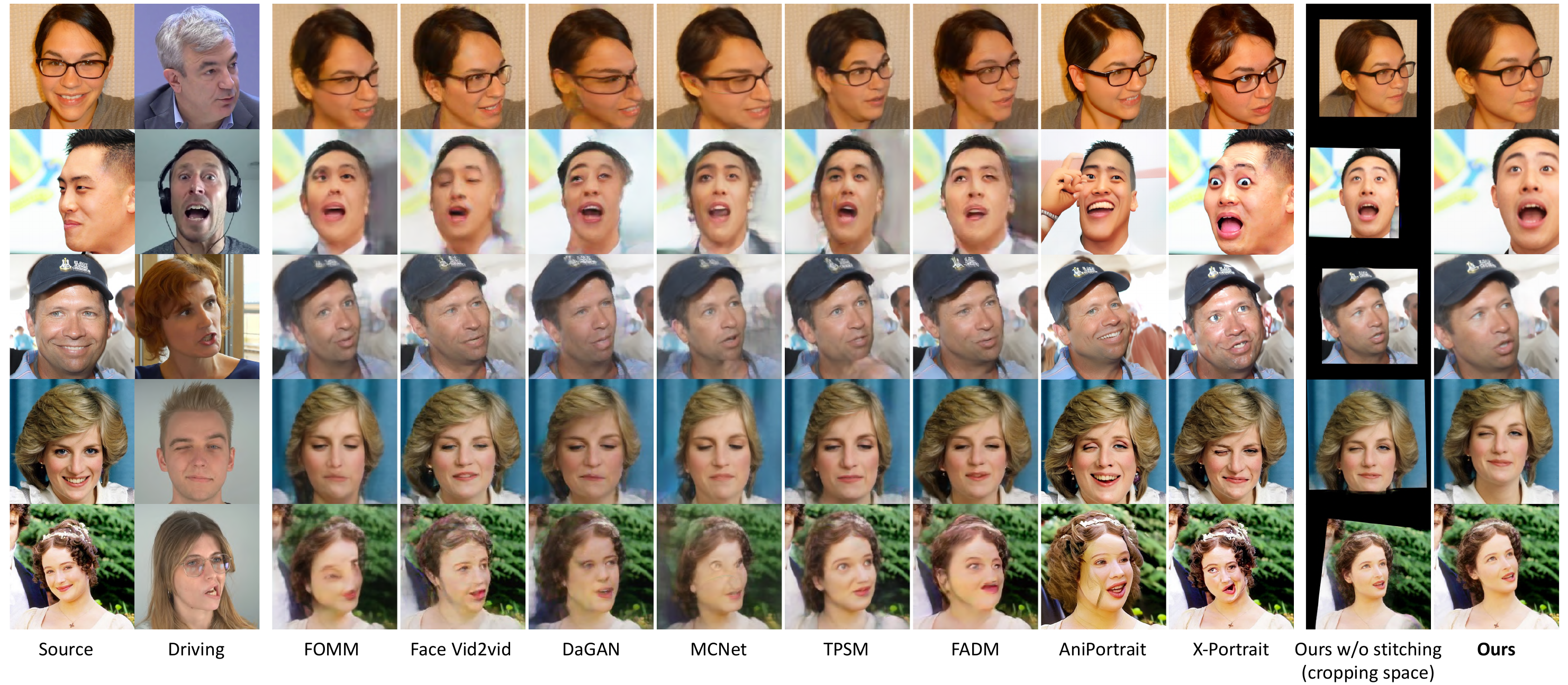}
  \caption{
    \textbf{Qualitative comparisons of cross-reenactment.}
    The first three source portraits are from FFHQ~\cite{karras2019style} and the last two are celebrities. Driving portraits are random selected from TalkingHead-1KH~\cite{wang2021facevid2vid}, VFHQ~\cite{xie2022vfhq} and NeRSemble~\cite{kirschstein2023nersemble}.
    We present the animated portraits without stitching in the cropping space, as well as the final results after stitching and pasting back into the original image space.
    Similar to self-reenactment, our model better transfers lip movements and eye gazes from another person, while maintaining the identity of the source portrait.}
  \label{fig:cross-driven}
\end{figure*}

\paragraph{Qualitative results.}
Qualitative comparisons of cross-reenactment are shown in \cref{fig:cross-driven}. The first two cases demonstrate our model's ability to stably transfer motion under large poses from the driving or source portraits. The third and fourth cases show that our model accurately transfers delicate lip movements and eye gazes, maintaining appearance details consistent with the source portrait. Additionally, the last case illustrates that stitching enables our model to perform stably even when the face region in the reference image is relatively small, providing the capability to animate multi-person inputs or full-body images gracefully.

\begin{table*}[t]
    \centering
    \resizebox{0.90\linewidth}{!}{
      \begin{tabular}{lcccccccccc}
        \toprule[1.5pt]
        \multirow{2}[2]{*}{Method}  & \multicolumn{5}{c}{\textbf{TalkingHead-1KH}} & \multicolumn{5}{c}{\textbf{VFHQ}} \\ \cmidrule[0.5pt](lr){2-6} \cmidrule[0.5pt](lr){7-11}
        & \!\!FID$\downarrow$\!\! & \!\!CSIM$\uparrow$\!\! & \!\!AED$\downarrow$\!\! & \!\!APD$\downarrow$\!\! & \!\!MAE (\textdegree)$\downarrow$\!\! & \!\!FID$\downarrow$\!\! & \!\!CSIM$\uparrow$\!\! & \!\!AED$\downarrow$\!\! & \!\!APD$\downarrow$\!\! & \!\!MAE (\textdegree)$\downarrow$\!\! \\
        \midrule[1pt]
         FOMM~\cite{siarohin2019first} & 90.8068 & 0.3057 & 0.7934 & 0.0411 & 18.3946 & 94.1640 & 0.2011 & 0.7374 & 0.0336 & 18.6282 \\
         Face Vid2vid~\cite{wang2021facevid2vid,facevid2vid3} & 82.9066 & 0.3687 & 0.8285 & 0.0559 & 20.2687 & 83.8891 & 0.2360 & 0.7891 & 0.0470 & 19.9852\\
         DaGAN~\cite{hong2022depth} & 81.1110 & 0.2937 & 0.7636 & 0.0405 & 21.0156 & 82.6255 & 0.1969 & 0.7108 & 0.0334 & 20.6918 \\
         MCNet~\cite{hong2023implicit} & 89.3218 & 0.2863 & \underline{0.7163} & \underline{0.0375} & \underline{17.0721} & 89.9694 & 0.1907 & \underline{0.6545} & 0.0329 & 17.3642 \\
         TPSM~\cite{zhao2022thin}  & 80.5436 & 0.3289 & 0.7492 & 0.0387 & 17.4371 & 77.5867 & 0.2197 & 0.6700 & \underline{0.0290} & \underline{16.8058} \\
         \midrule[1pt]
         FADM~\cite{zeng2023face} & 95.4043 & \underline{0.3755} & 0.8158 & 0.0525 & 18.8346 & 98.2516 & 0.2473 & 0.7811 & 0.0438 & 18.9776 \\
         AniPortrait~\cite{wei2024aniportrait} & \textbf{47.8739} & 0.3733 & 0.9127 & 0.0450 & 19.7136 & \underline{70.8077} & \underline{0.2538} & 0.9018  & 0.0501 & 20.1085 \\
         X-Portrait~\cite{xie2024x} & 60.7963 & \textbf{0.5843} & 0.8392 & 0.1070 & 20.9344 & \underline{58.6731} & \textbf{0.5881} & 0.8463  & 0.1226 & 22.5937 \\
         \midrule[1pt]
        \textbf{Ours} & \underline{58.0370} & \underline{0.3909} & \textbf{0.6772} & \textbf{0.0333} & \textbf{14.7946} & \textbf{56.4165} & \underline{0.2606} & \textbf{0.6476} & \textbf{0.0271} & \textbf{13.3464} \\ 
        \bottomrule[1.5pt]
    \end{tabular}
    }
    \caption{\textbf{Quantitative comparisons of cross-reenactment.}}
    \label{tab:cross-driven}
\end{table*}

\begin{figure*}[htpb]
  \centering
  \includegraphics[width=0.95\linewidth]{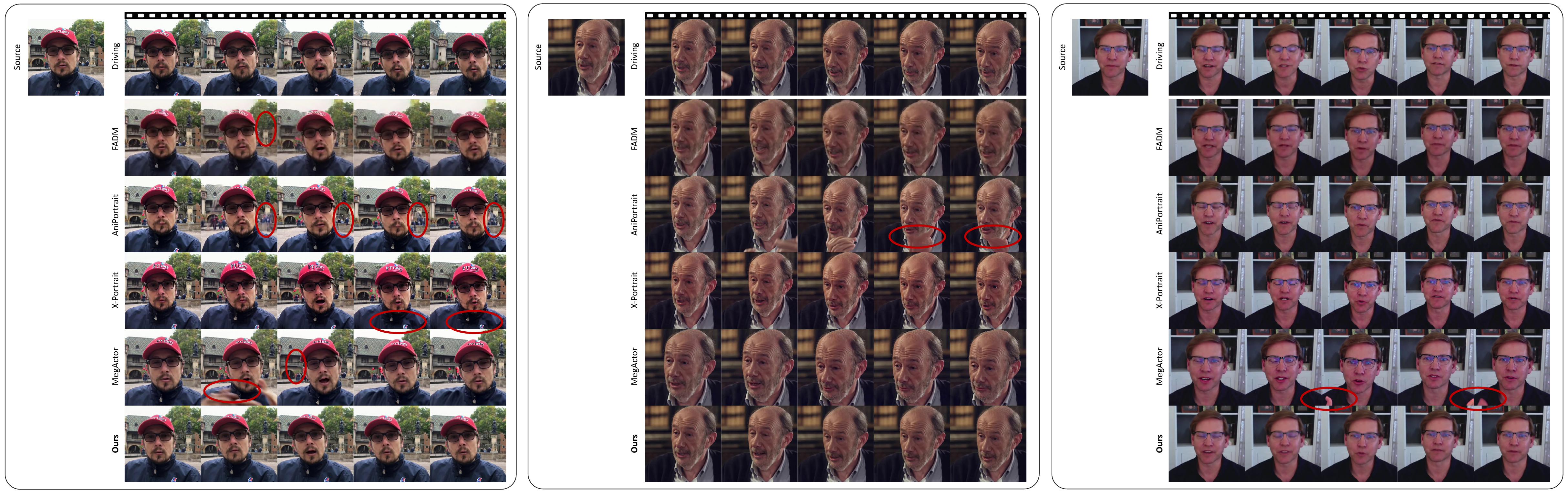}
  \caption{
    \textbf{Temporal consistency comparisons with diffusion-based methods.}
    These three cases are from VFHQ and TalkingHead-1KH test sets.
    Our animation results are in the original image space with stitching.
    Within the vertical circles, the statue disappears in the subsequent animated frames of FADM~\cite{zeng2023face}, there are pedestrian-like unnatural background movements in the animated results of AniPortrait~\cite{wei2024aniportrait}, and the red banner disappears in some frames of MegActor~\cite{yang2024megactor}. Within the horizontal circles, there are hand-waving-like unnatural foreground movements in the animated images of both AniPortrait~\cite{wei2024aniportrait} and MegActor~\cite{yang2024megactor}, while the patterns on the clothing change in the animated images of X-Portrait~\cite{xie2024x}.
  }
  \label{fig:stableness}
\end{figure*}

\paragraph{Quantitative results.}
\cref{tab:cross-driven} shows the quantitative results of the cross-reenactment comparisons. Our model outperforms previous diffusion-based and non-diffusion-based methods in both generation quality and motion accuracy, except for the FID on TalkingHead-1KH and the CSIM on both datasets, where the FID of the diffusion-based method AniPortrait~\cite{wei2024aniportrait} and the CSIM of X-Portrait~\cite{xie2024x} are better than ours.
The flip side is that diffusion-based methods require much more inference time than non-diffusion-based methods due to multiple denoising steps and high FLOPs. Additionally, the temporal consistency of the foreground and background is not as good compared to non-diffusion-based methods, due to the high variability of the diffusion models.
This phenomenon can be observed in \cref{fig:stableness}, where self-reenactment cases from testsets of VFHQ and TalkingHead-1KH are exemplified. We also illustrate the results of MegActor~\cite{yang2024megactor} in \cref{fig:stableness}, which is also a diffusion-based method sharing a similar mutual self-attention and plugged temporal attention architecture as AniPortrait~\cite{wei2024aniportrait} and AnimateAnyone~\cite{hu2023animate}.

\subsection{Ablation Study and Analysis}
In this section, we discuss the benefits and necessities of stitching, eyes and lip retargeting.

\paragraph{Ablation of the stitching module.}
\begin{figure*}[htpb]
  \centering
  \includegraphics[width=0.9\linewidth]{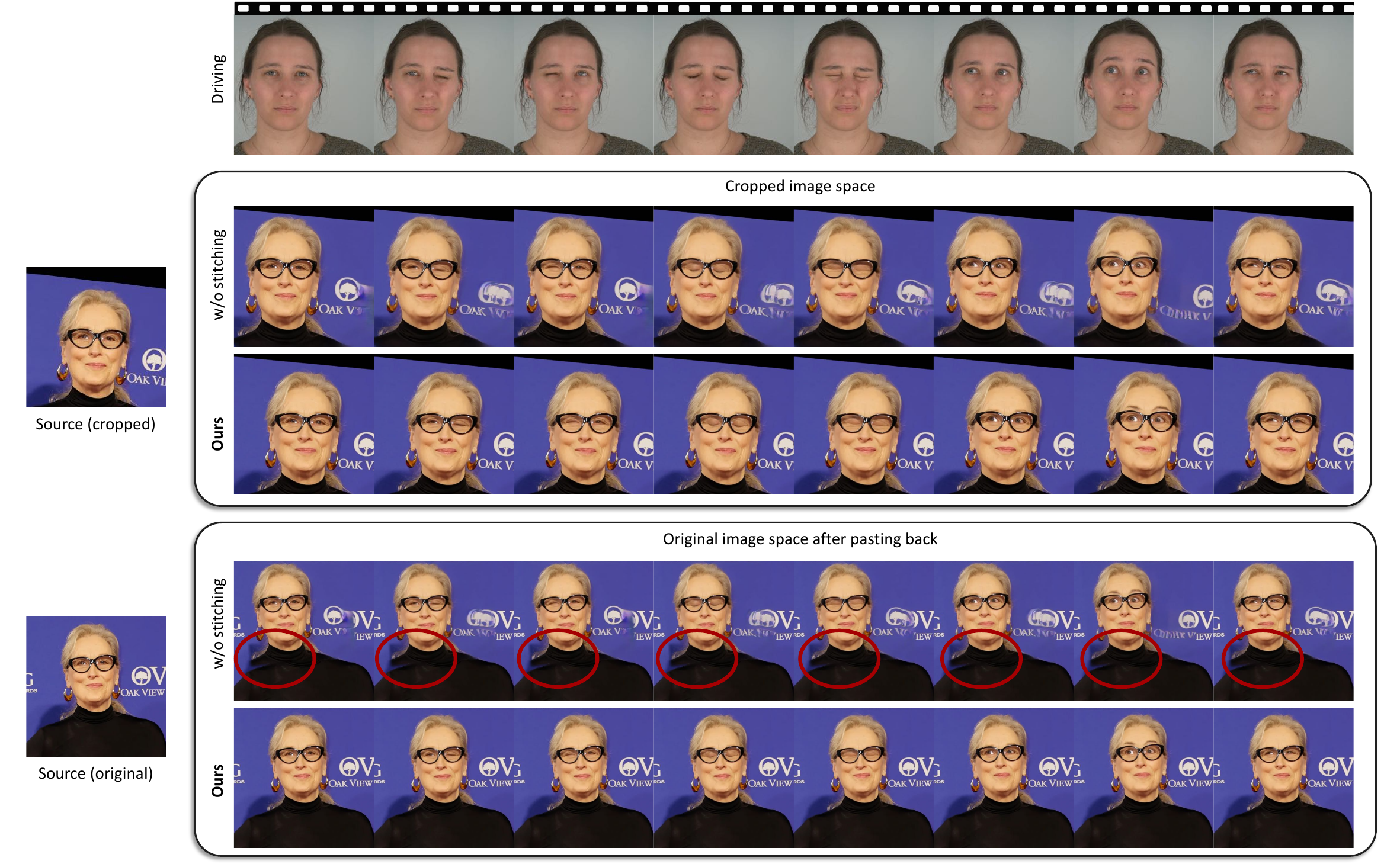}
  \caption{
    \textbf{Ablation study of the stitching.}
    The first block shows the comparisons of stitching in the cropping image space, and the second block shows the comparisons after mapping into the original image space. The misalignment is apparent without stitching, especially in the shoulder region.
  }
  %
  %
  %
  %
  %
  \label{fig:shoulder_retargeting}
\end{figure*}
%
%

%
As shown in the first block of \cref{fig:shoulder_retargeting}, given a source image and a driving video sequence, the animated results without stitching share the same shoulder position as the driving frames.
After stitching, the shoulder of the animated person is force aligned with the cropped source portrait while preserving the motion and appearance.
In the second block of \cref{fig:shoulder_retargeting}, after mapping into the original image space, it is clear that the animated portrait without stitching shows significant shoulder misalignment, while the stitched results show no visually apparent misalignment.

\paragraph{Ablation of eyes retargeting.}
\begin{figure*}[htpb]
  \centering
  \includegraphics[width=0.92\linewidth]{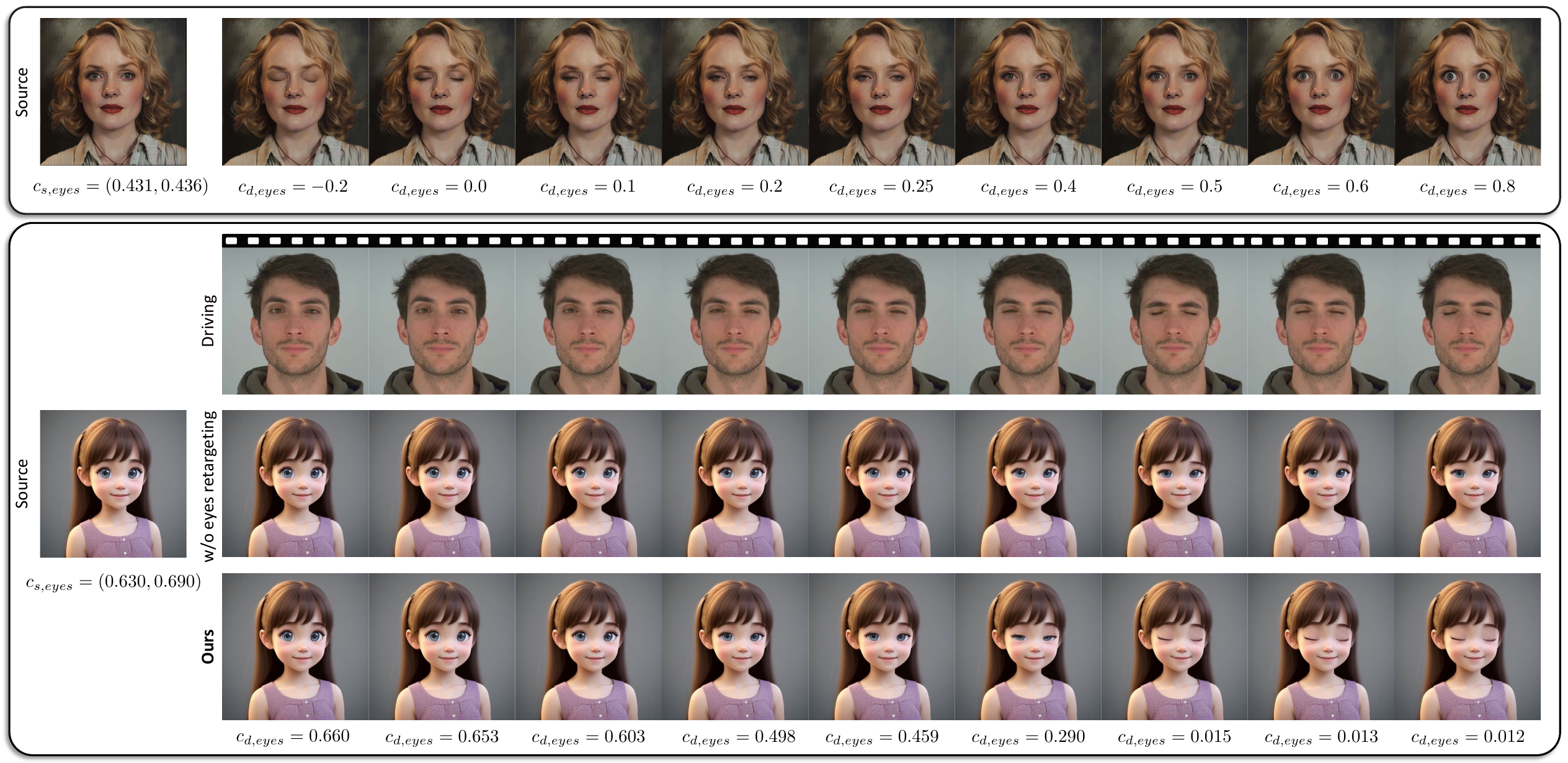}
  \caption{\textbf{Examples and ablation study of our eyes retargeting.}
    The first block shows the eyes-open controllability of our model on the source image without any driving frames.
    The second block demonstrates the ability of eye retargeting in cross-reenactment, especially when the eyes of the source person are much larger than the driving one.
    For clarity, the animated results adopt the source head rotation.
  }
  \label{fig:eye_retargeting}
\end{figure*}
In the first block of \cref{fig:eye_retargeting}, the quantitative controllability of the eyes-open of the source image is illustrated.
Without any driving motions, one can provide a proper driving eyes-open scalar $c_{d,eyes}$ from 0 to 0.8, send it to the eyes retargeting module $\mathcal{R}_{eyes}$ along with the source eyes-open condition tuple $c_{s,eyes}$, and drive the eyes from closed to fully open.
The eyes-open motion does not affect the remaining part of the reference image.
Additionally, an out-of-training-distribution driving eyes-open scalar, such as $-0.2$, can also achieve reasonable results.
%
%
%
%
In the second block of \cref{fig:eye_retargeting}, a cartoon image is animated by the driving frames of a closing-eye video.
The eyes of the girl are much larger than those of the man in the first driving frame.
Therefore, the driving eye-closing motion is too weak to close the girl's eyes to the same extent, as observed in the second row without eyes retargeting.
When we employed eyes retargeting, the driving eyes-open scalar $c_{d,eyes,i}$ corresponding to the $i$-th driving frame can be formulated as:
$
  c_{d,eyes,i} = \overline{c}_{s,eyes} \cdot \frac{\overline{c'}_{s,eyes,i}}{\overline{c'}_{s,eyes,0}},
$
where $\overline{c'}_{s,eyes,i}$ is the average value of the eyes-open condition tuple for the $i$-th driving frame, and the overline represents the averaging operation.
Benefiting from eyes retargeting, the animated frames achieve the same eye-closing motion as the driving video.

%

%
\begin{figure*}[htpb]
  \centering
  \includegraphics[width=0.95\linewidth]{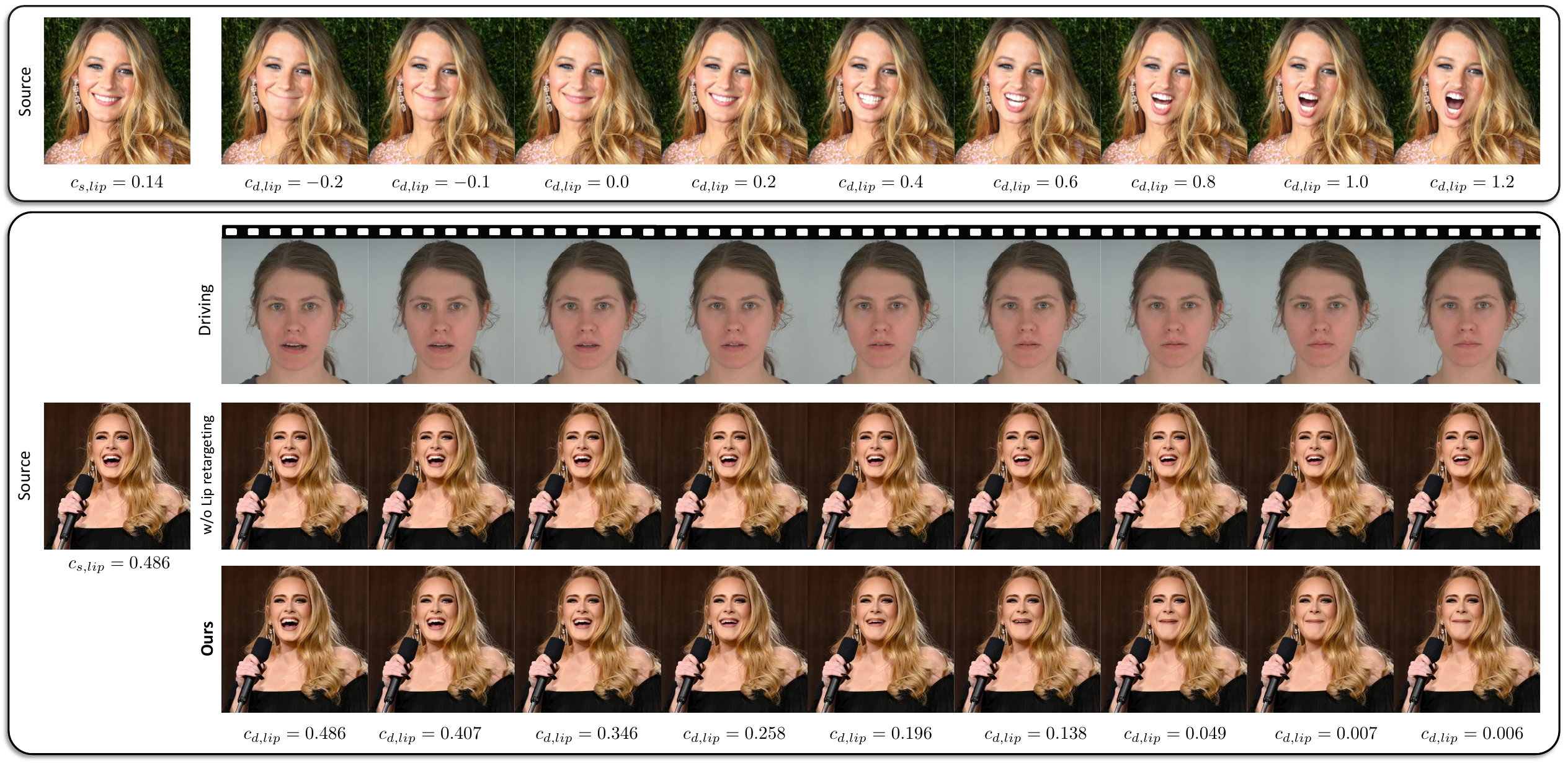}
  \caption{
    \textbf{Examples and ablation study of the lip retargeting.}
    Similar to eye retargeting, these two blocks show our controllability conditioned on arbitrary lip-open scalars, either randomly sampled or extracted from driving frames.
  %
  }
  \label{fig:mouth_retargeting}
\end{figure*}
\paragraph{Ablation of lip retargeting.}
Similarly, the first block of \cref{fig:mouth_retargeting} illustrates the quantitative lip-open controllability of the source image.
One can input a driving lip-open scalar $c_{d,lip}$ between 0 and 0.8, feed it to the lip retargeting module along with the source lip-open condition $c_{s,lip}$, and drive the lips from closed to fully open.
The lip-open motion does not affect the remaining part of the source image.
%
%
%
%
%
An out-of-training-distribution driving lip-open scalar can also achieve reasonable results, as depicted in \cref{fig:mouth_retargeting}.
Additionally, the tongue is generated when the lips are widely open.
As shown in the second block of \cref{fig:mouth_retargeting}, we can also drive the lip to close conditioned on a lip-close scalar $c_{d,lip,i}$ extracted from the driving frame:
$
  c_{d,lip,i} = c_{s,lip} \cdot \frac{c'_{s,lip,i}}{c'_{s,lip,0}},
$
where $c'_{s,lip,i}$ is the lip-open condition of the $i$-th driving frame.

%
%
%
%
\begin{figure*}[htpb]
  \centering
  \includegraphics[width=0.9\linewidth]{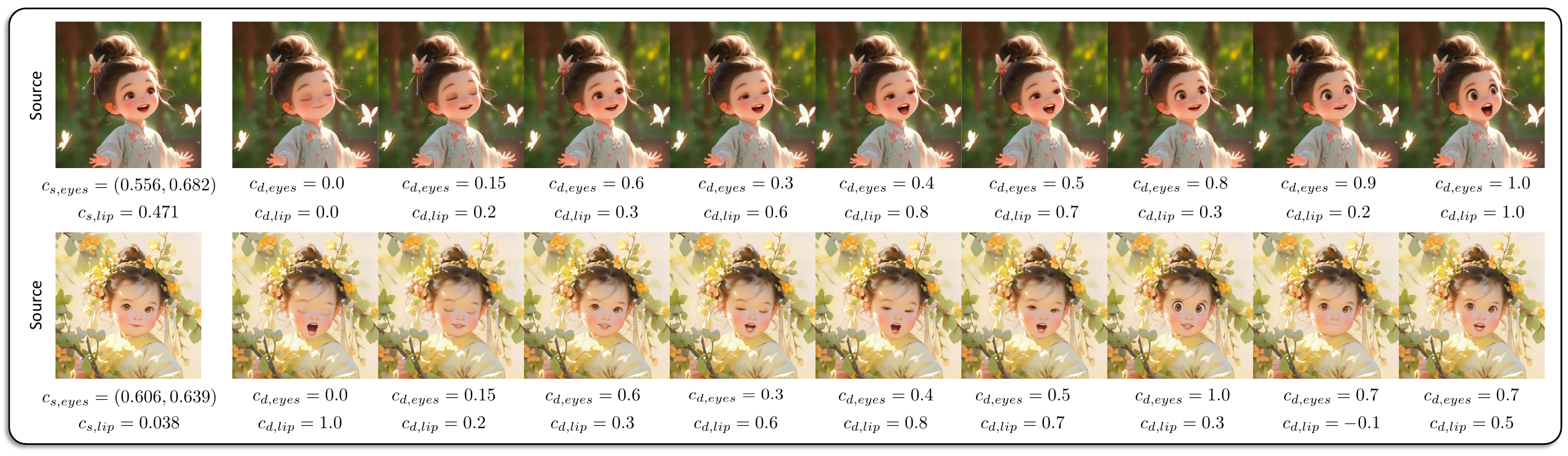}
  \caption{\textbf{Examples of simultaneous eyes and lip retargeting.}
    Given driving eyes-open and lip-open scalars simultaneously, the animated results from the source image suggest that eye and lip retargeting can be effective simultaneously, even though these two retargeting modules are trained independently.
    %
    }
  \label{fig:mouth_eye_retargeting}
\end{figure*}
\paragraph{Analysis on eyes and lip retargeting.}
%
%
%
A natural question is whether the eye and lip retargeting can take effect simultaneously, as described in \cref{algo:inference}, when \(\alpha_{eyes} = 1\) and \(\alpha_{lip} = 1\).
In other words, the core question is whether the retargeting modules have learned to distinguish the different patterns of $\Delta_{eyes}$ and $\Delta_{lip}$.
The two animation examples in \cref{fig:mouth_eye_retargeting} positively support this hypothesis.
The source images of the girls in ancient costumes can be animated to reasonable results with the given driving eyes-open and lip-open scalars, where $\Delta_{eyes}$ and $\Delta_{lip}$ are added to the driving keypoints simultaneously.
%
%
%

\section{Conclusion}
\label{sec:conclusion}

In this paper, we present an innovative video-driven framework for animating static portrait images, making them realistic and expressive while ensuring high inference efficiency and precise controllability.
The generation speed of our model achieves 12.8ms on an RTX 4090 GPU using the naive Pytorch framework, while simultaneously outperforming other heavy diffusion-based methods.
We hope our promising results pave the way for real-time portrait animation applications in various scenarios, such as video conferencing, social media, and entertainment, as well as audio-driven character animations.

\paragraph{Limitations.} Our current model struggles to perform well in cross-reenactment scenarios involving large pose variations. Additionally, when the driving video involves significant shoulder movements, there is a certain probability of resulting in jitter. We plan to address these limitations in future work.

\paragraph{Ethics considerations.} Portrait animation technologies pose social risks, including misuse for deepfakes. To mitigate these risks, ethical guidelines and responsible use practices are essential. Currently, the synthesized results exhibit some visual artifacts that could aid in detecting deepfakes.

\section*{Acknowledgments}
We would like to thank our colleagues and partners Haotian Yang, Haoxian Zhang, Mingwu Zheng, Chongyang Ma, and others for their valuable discussions and insightful suggestions on this work.

{
    \small
    \bibliographystyle{unsrt} 
    \bibliography{11_references}
}

\clearpage \appendix 
\section{Benchmark Metric Details}
\label{sec:appendix_benchmark}
\paragraph{LPIPS.} We use the AlexNet based perceptual similarity metric LPIPS~\cite{zhang2018unreasonable} to measure the perceptual similarity between the animated and the driving images.

\paragraph{AED.} AED is the mean $\mathcal{L}_1$ distance of the expression parameters between the animated and the driving images. These parameters, which include facial movement, eyelid, and jaw pose parameters, are extracted by the state-of-the-art 3D face reconstruction method SMIRK~\cite{retsinas20243d}.


\paragraph{APD.} APD is the mean $\mathcal{L}_1$ distance of the pose parameters between the animated and the driving images. The pose parameters are extracted by SMIRK~\cite{retsinas20243d}.

\paragraph{MAE.} To measure the eyeball direction error between the animated and the driving images, the mean angular error (\textdegree) is adopted as:
%
$
    \textrm{MAE}(I_{p}, I_{d}) = \textrm{arccos}(\frac{\textbf{b}_p \cdot \textbf{b}_d}{\Vert \textbf{b}_p \Vert \cdot \Vert \textbf{b}_d \Vert}),
$
%
where $\textbf{b}_p$ and $\textbf{b}_d$ are the eyeball direction vectors of the animated image $I_{p}$ and the driving image $I_{d}$ respectively, and they are predicted by a pretrained eyeball direction network~\cite{abdelrahman2023l2cs}.

\paragraph{CSIM.} CSIM measures the identity preservation between two images, through the cosine similarity of two embeddings from a pretrained face recognition network~\cite{deng2019arcface}. For self-reenactment, the CSIM is calculated between the animated and the driving images. For cross-reenactment, the CSIM is calculated between the animated and the source portraits.

%

\paragraph{FID.} FID compares the distribution of animated images with the distribution of a set of real images. For TalkingHead-1KH test set, FID is calculated between the animated images and the last 38,400 images of the FFHQ dataset. For the VFHQ test set, FID is calculated between the animated images and the last 15,000 images of FFHQ.

%
%
%
%

\paragraph{Dataset processing.}
We follow~\cite{facevid2vid-set} to pre-process the evaluation set of the TalkingHead-1KH. In cross-reenactment, we extract 1 frame every 10 frames from each video, for a total of 24 frames as the driving sequence.
For VFHQ, we extract 1 frame every 5 frames, for a total of 6 frames.
It is observed that insufficient driving video length might reduce the animation quality of X-Portrait~\cite{xie2024x}. 
Therefore, in the cross-reenactment experiments for X-Portrait, we use the first 250 frames of each video in the TalkingHead-1KH test set and the full videos in the VFHQ test set as the driving videos. 
Subsequently, we extract frames from the animated results that match the indices of the comparison frames used by other methods.

\begin{figure}[htpb]
    \centering
    \includegraphics[width=0.975\linewidth]{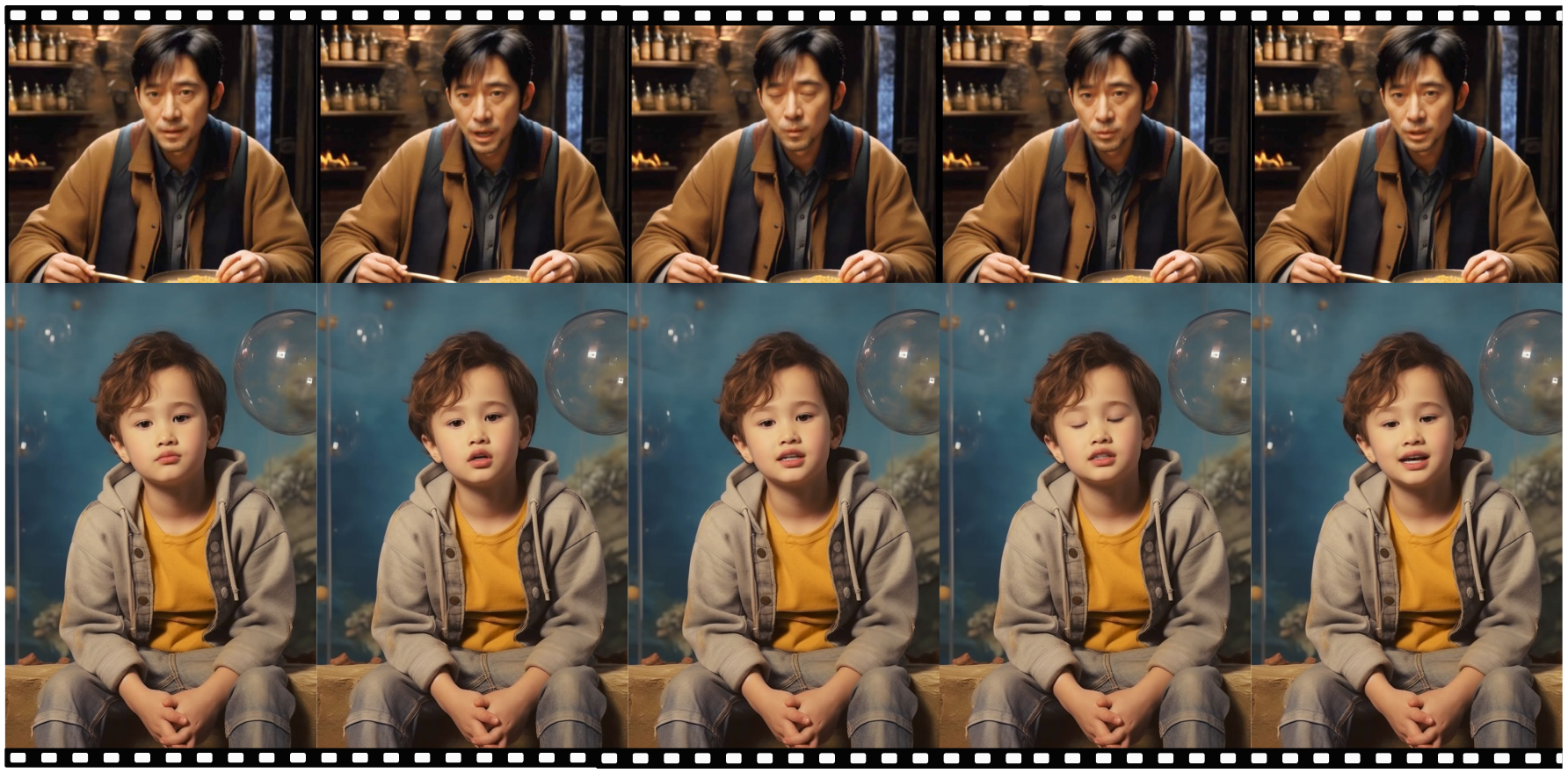}
    \caption{
        \textbf{Audio-driven examples.} This figure presents two examples of audio-driven portrait animation with stitching applied. The lip movements can be accurately driven by the audios input.
    }
    \label{fig:audio_driven}
\end{figure}

\begin{figure}[htpb]
    \centering
    \includegraphics[width=0.99\linewidth]{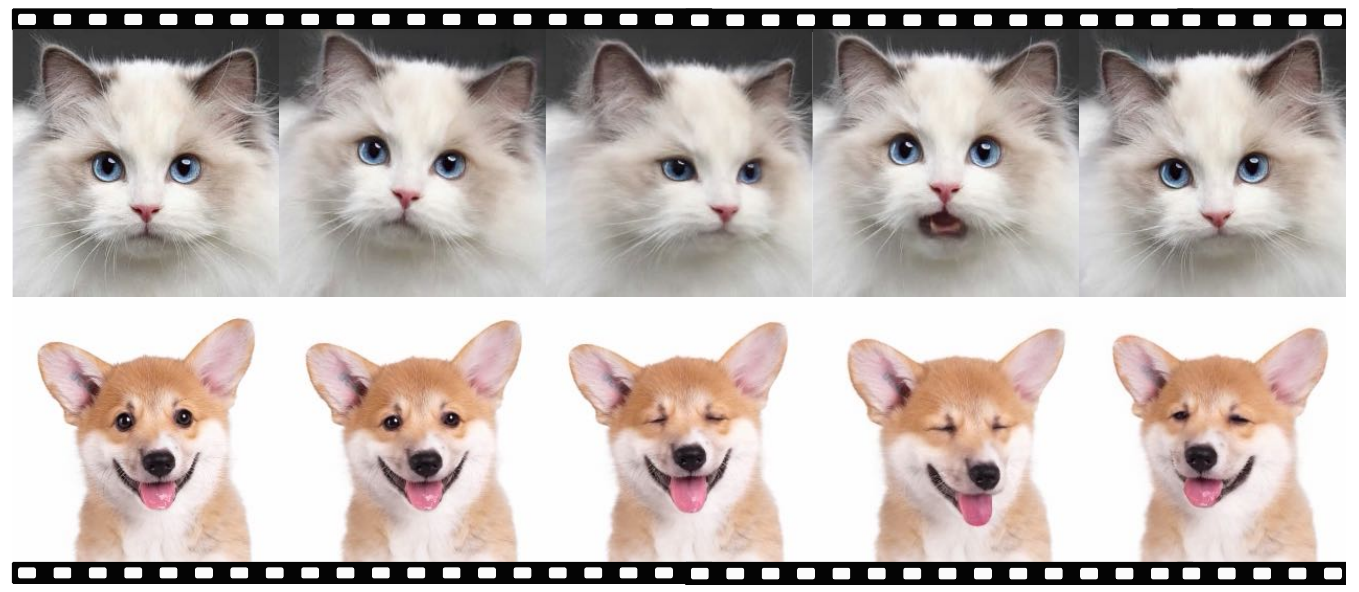}
    \caption{
        \textbf{Animal animation examples.} We show the animation results of a Ragdoll cat and a Corgi dog, with driving motions derived from human videos.
    }
    \label{fig:gen2animal}
\end{figure}

\section{Qualitative Results on Multi-person Portrait}
\begin{figure*}[htpb]
    \centering
    \includegraphics[width=0.955\linewidth]{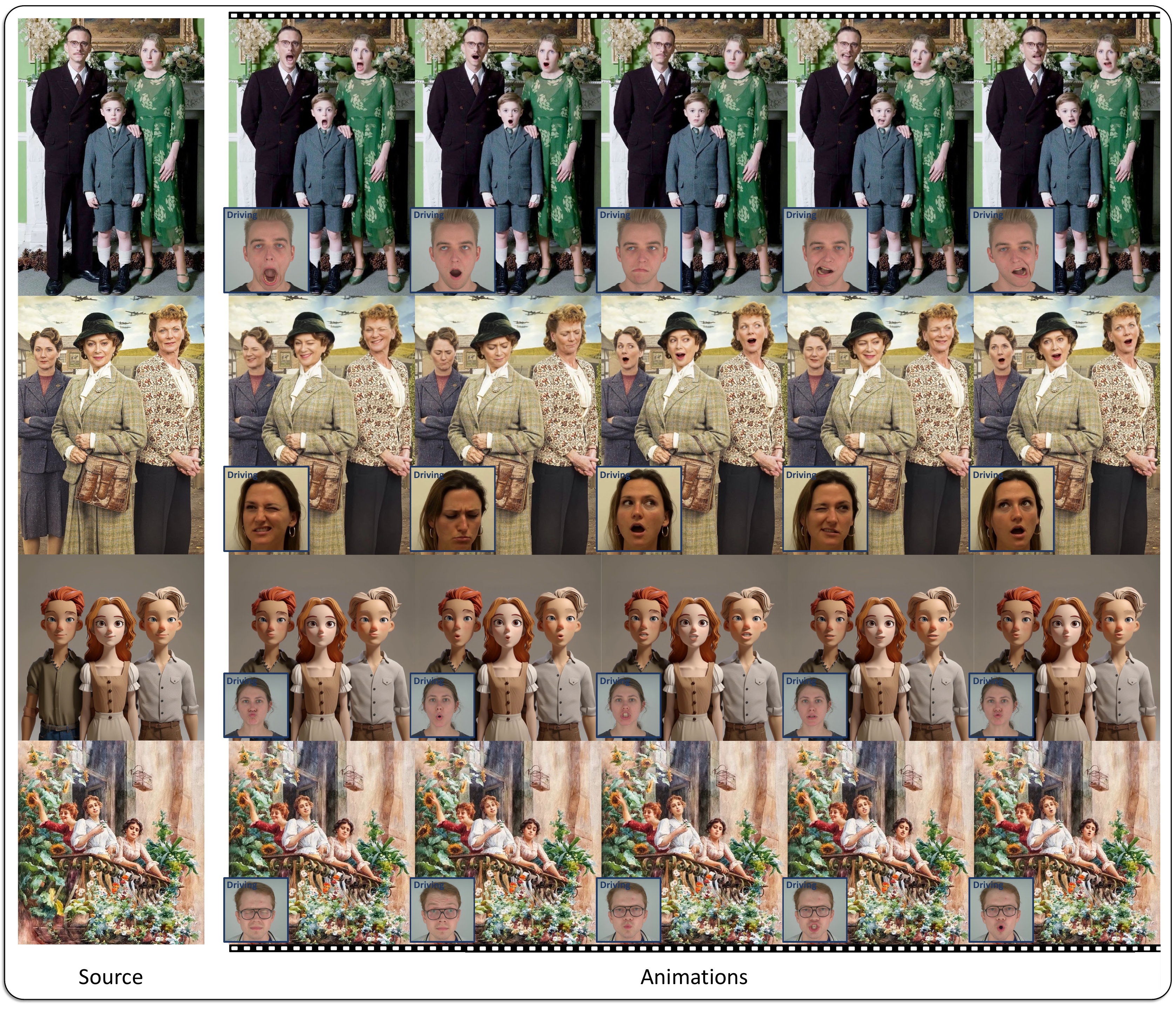}
    \caption{
        \textbf{Multi-person portrait animation examples.} Given a group photo of several subjects and a driving video sequence, our model can animate each subject with the stitching applied. The driving frame corresponding to each animated image is located in the left-down corner of the animated image.
    }
    \label{fig:multi_person}
\end{figure*}
We provide additional qualitative results on multi-person portrait animation in \cref{fig:multi_person}.
Benefiting from the stitching ability of our model, each person in the portrait can be animated separately.

\section{Audio-driven Portrait Animation}
We can easily extend our video-driven model to audio-driven portrait animation by regressing or generating motions, including expression deformations and head poses, from audio inputs.
For instance, we use Whisper~\cite{radford2022whisper} to encode audio into sequential features and adopt a transformer-based framework, following FaceFormer~\cite{faceformer2022}, to autoregress the motions.
The audio-driven results are shown in \cref{fig:audio_driven}.

\section{Generalization to Animals}
We find our model can generalize well to animals, \eg, cats and dogs, by fine-tuning on a small dataset of animal portraits combined with the original data.
Specifically, in the fine-tuning stage, we discard the head pose loss term $\mathcal{L}_H$, the lip GAN loss, and the faceid loss $\mathcal{L}_{\text{faceid}}$, as the head poses of animals are not as accurate as those of humans, the lip distribution differs from humans, and faceid cannot be applied to animals.
Surprisingly, we can drive the animals with human driving videos, and the results are shown in \cref{fig:gen2animal}.

\section{Portrait Video Editing}
\begin{figure*}[htpb]
    \centering
    \includegraphics[width=0.975\linewidth]{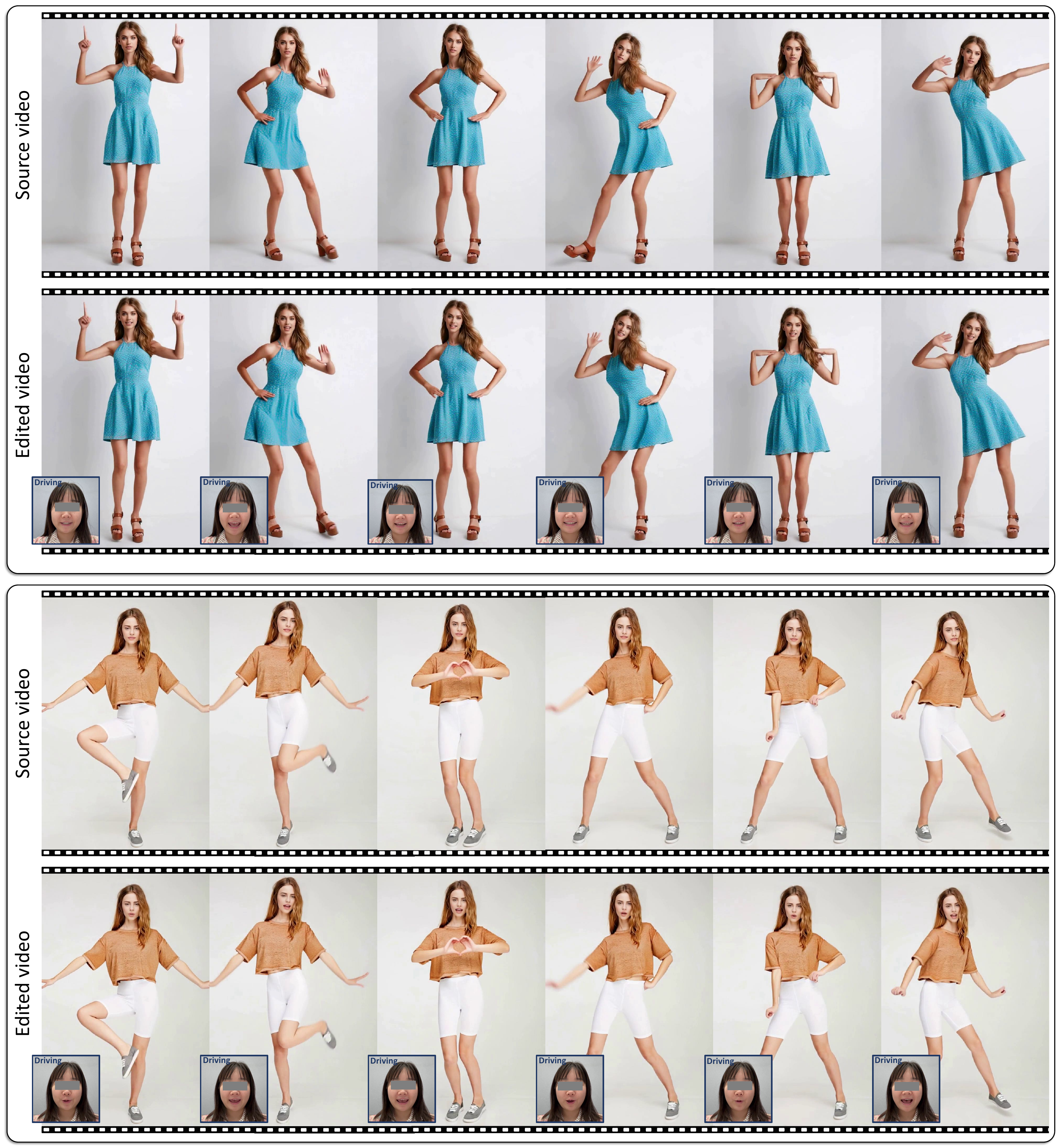}
    \caption{
        \textbf{Portrait video editing examples.} Given a source video sequence, such as a dancing video, our model can re-animate the head part using a driving video sequence. The edited video frame inherits the expression from the driving frame while preserving the non-head regions from the source frame.
    }
    \label{fig:portrait-video}
\end{figure*}

We can extend our model to edit the head region of a source video sequence, while minimally sacrificing the temporal consistency of the source video.
The source and driving implicit keypoints in Eqn.~\ref{eqn:inference} are transformed as follows:
\begin{equation}
    \left\{
        \begin{array}{ll}
        x_{s,i} &= s_{s,i} \cdot (x_{c,s,i} R_{s,i} + \delta_{s,i}) + t_{s,i}, \\
        x_{d,i} &= s_{s,i} \cdot \frac{s_{d,i}}{s_{d,0}} \cdot \big( x_{c,s,i} (R_{d,i} R^{-1}_{d,0} R_{s,i}) + \\
                        &(\delta_{s,i} + 0.5 \cdot (\delta_{d,i} + \delta_{d,i+1}) - \delta_{d,0}) \big) + \\
                        & (t_{s,i} + t_{d,i} - t_{d,0}),
        \end{array}
    \right.
\end{equation}
where $x_{s,i}, s_{s,i}, x_{c,s,i}, R_{s,i}, \delta_{s,i}, t_{s,i}$ represent the source keypoints, scale factor, canonical implicit keypoints, head pose, expression deformation, and translation of the $i$-th source frame, respectively. The operation $0.5 \cdot (\delta_{d,i} + \delta_{d,i+1})$ performs smoothing by averaging the expression offsets of the $i$-th and $(i+1)$-th driving frames. As exemplified in \cref{fig:portrait-video}, the edited frame with stitching applied inherits the expression from the corresponding driving frame, while preserving the non-head regions from the corresponding source frame.


\end{document}